\documentclass[sn-basic,iicol,Numbered]{sn-jnl}

\usepackage{graphicx}%
\usepackage[export]{adjustbox}
\usepackage{multirow}%
\usepackage{amsmath,amssymb,amsfonts}%
\usepackage{amsthm}%
\usepackage{mathrsfs}%
\usepackage[title]{appendix}%
\usepackage{xcolor}%
\usepackage{textcomp}%
\usepackage{manyfoot}%
\usepackage{booktabs}%
\usepackage{algorithm}%
\usepackage{algorithmicx}%
\usepackage{algpseudocode}%
\usepackage{listings}%
\usepackage[]{xcolor}
\usepackage[normalem]{ulem}
\useunder{\uline}{\ul}{}

\usepackage{lipsum}

\newcommand\blfootnote[1]{%
  \begingroup
  \renewcommand\thefootnote{}\footnote{#1}%
  \addtocounter{footnote}{-1}%
  \endgroup
}

\definecolor{greenF2}{HTML}{00c41e}
\definecolor{redF2}{HTML}{FF5959}
\definecolor{blueF2}{HTML}{5959FF}
\definecolor{blueL}{HTML}{95aaff}
\definecolor{orangeF2}{HTML}{ffa000}
\definecolor{magentaF2}{HTML}{FF7FFF}
\definecolor{yellowF2}{HTML}{ffdd37}
\definecolor{oilF2}{HTML}{d2dd37}
\definecolor{purpleF2}{HTML}{E0B0FF}
\definecolor{cyanF2}{HTML}{00d4f7}
\definecolor{green2F2}{HTML}{00e884}
\definecolor{beigeF2}{HTML}{fff795}

\theoremstyle{thmstyleone}%
\theoremstyle{thmstyletwo}%

\theoremstyle{thmstylethree}%

\def\ie{\emph{i.e.}}
\def\etc{\emph{etc.}}

\begin{document}

\title[Look Into the LITE]{Look Into the LITE in Deep Learning for Time Series Classification}


\author*[1]{\fnm{Ali} \sur{Ismail-Fawaz}}\email{ali-el-hadi.ismail-fawaz@uha.fr}

\author[1]{\fnm{Maxime} \sur{Devanne}}\email{maxime.devanne@uha.fr}

\author[2]{\fnm{Stefano} \sur{Berretti}}\email{stefano.berretti@unifi.it}

\author[1]{\fnm{Jonathan} \sur{Weber}}\email{jonathan.weber@uha.fr}

\author[1,3]{\fnm{Germain} \sur{Forestier}}\email{germain.forestier@uha.fr}

\affil*[1]{\orgdiv{IRIMAS}, \orgname{Universite de Haute-Alsace}, \orgaddress{\city{Mulhouse}, \country{France}}}

\affil[2]{\orgdiv{MICC}, \orgname{University of Florence}, \orgaddress{\city{Florence}, \country{Italy}}}

\affil[3]{\orgdiv{DSAI}, \orgname{Monash University}, \orgaddress{\city{Melbourne}, \country{Australia}}}

\abstract{
Deep learning models have been shown to be a powerful solution for Time Series Classification (TSC). State-of-the-art architectures, while producing promising results on the UCR
and the UEA archives
, present a high number of trainable parameters.
This can lead to long training with high CO2 emission, power consumption and possible increase in the number of FLoating-point Operation Per Second (FLOPS).
In this paper, we present a new architecture for TSC, the \textbf{Light Inception with boosTing tEchnique (LITE)} with only $2.34\%$ of the number of parameters of the state-of-the-art InceptionTime model, while preserving performance.
This architecture, with only $9,814$ trainable parameters due to the usage of DepthWise Separable Convolutions (DWSC), is boosted by three techniques: multiplexing, custom filters, and dilated convolution. 
The LITE architecture, trained on the UCR, is $2.78$ times faster than InceptionTime and consumes $2.79$ times less CO2 and power, while achieving an average accuracy of $84.62\%$ compared to $84.91\%$ with InceptionTime.
To evaluate the performance of the proposed architecture on multivariate time series data, we adapt LITE to handle multivariate time series, we call this version LITEMV.
To bring theory into application, we also conducted experiments using LITEMV on multivariate time series representing human rehabilitation movements, showing that LITEMV not only is the most efficient model but also the best performing for this application on the Kimore dataset, a skeleton based human rehabilitation exercises dataset.
Moreover, to address the interpretability of LITEMV, we present a study
using Class Activation Maps to understand the classification decision taken by the model during evaluation.
}




\maketitle
\blfootnote{This is the author’s version of an article published in the International Journal of Data Science and Analytics.
The final authenticated version is available online at: \url{https://doi.org/10.1007/s41060-024-00708-5}}

\section{Introduction}\label{sec1}

\begin{figure}
    \centering
    \includegraphics[width=\linewidth]{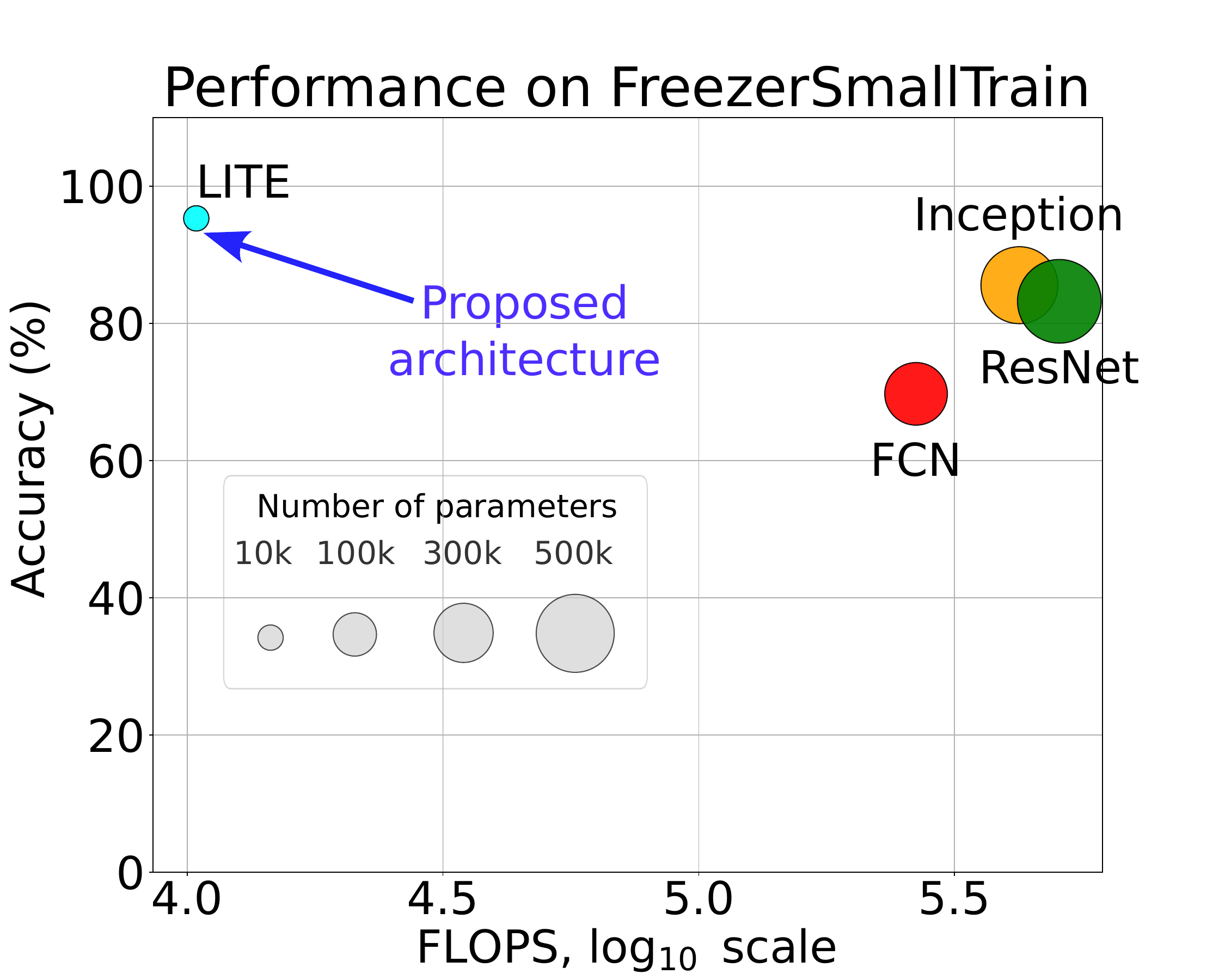}
    \caption{For each model, the accuracy on the FreezerSmallTrain dataset is presented on the $y$-axis and the number of FLoat-point Operations Per Second (FLOPS) is presented on the $x$-axis in $\log_{10}$ scale.
    The diameter of the circles represents the number of trainable parameters of the model.
    The smallest model is LITE (\textbf{ours}) with only $~10k$ trainable parameters and the lowest number of FLOPS ($~4$ in $\log_{10}$ scale); it also presents the highest accuracy score on the test set in this comparison.}
    \label{fig:summary}
\end{figure}

Time Series Classification (TSC) has been widely investigated by researchers in recent years.
Some TSC tasks include the classification of surgical evaluation~\cite{ismail2018evaluating,tao2012sparse,forestier2012classification}, action recognition of human motion~\cite{devanne20143,ji20123d}, cheat detection in video games~\cite{pinto2021deep}, interpretability~\cite{younis2022multivariate}, Entomology~\cite{madrid2019matrix}, \etc~
Thanks to the availability of the UCR archive~\cite{dau2019ucr}, 
and the UEA archive~\cite{bagnall2018uea}, the largest archives for univariate and multivariate
TSC datasets, a significant amount of work has been done in this domain.
Deep learning models have been proposed in the time series context for classification~\cite{ismail2019deep,fcn_resnet,ismail2020inceptiontime,ismail-fawaz2022hccf,pialla2022datsc}, clustering~\cite{lafabregue2022end}, averaging~\cite{terefe2020time}, representation learning~\cite{ismail-fawaz2022trilite,zerveas2021representation,mixingUp}, adversarial attacks~\cite{fawaz2019adversarial,pialla2022smooth}, \etc~
Even though deep learning approaches proven to be very powerful for TSC, they present a large amount of trainable parameters, which often leads to a long training time, inference time and storage usage.

For this reason, some works started to question the need of such a large complexity in deep learning models for TSC such as ROCKET and its variants~\cite{dempster2020rocket,dempster2021minirocket,tan2022multirocket}.
Like for images, deep learning also presents a large complexity, which limits the usage of the models on small devices such as mobile phones and robots~\cite{howard2017mobilenets,sandler2018mobilenetv2}.
Furthermore, Large Language Models (LLM) also shown to be very effective~\cite{brown2020language}, and that their complexity can be decreased, while preserving performance~\cite{schick2020s}.

In this paper, we address the methodology of reducing the complexity of deep learning models, while preserving the performance of the TSC task.
We argue that a large complex model may not be necessary in order to perform well on the UCR archive.
However, simply removing layers and or parameters to reduce complexity may not guarantee to preserve the performance.
For this reason, the neural network architecture often requires additional techniques in order to balance between complexity and performance.
In this work, we borrow existing techniques that have been efficiently used in state-of-the-art architectures on time series data.
These techniques are multiplexing convolutions~\cite{ismail2020inceptiontime,cui2016multi}, dilated convolutions~\cite{mixingUp}, and custom filters~\cite{ismail-fawaz2022hccf}.
By combining these three techniques with a modified version of a small non complex model, the Fully Convolution Network (FCN)~\cite{fcn_resnet}, we propose a new architecture, named Light Inception with boosTing tEchniques (\textbf{LITE}).
The proposed model uses only $2.34\%$ of the number of parameters of the Inception model, while being competitive with state-of-the-art architectures.
For instance, Figure~\ref{fig:summary} shows that on the FreezerSmallTrain dataset, the classification accuracy of \textbf{LITE} is much higher than other approaches with way less trainable parameters.
The reduction in number of parameters is
made possible thanks to the usage of DepthWise Separable Convolutions (DWSC)~\cite{howard2017mobilenets}.
The additional techniques used in this proposed architecture, multiplexing, dilated, and custom convolutions,
have the advantage of only slightly increasing the number of parameters by about $1,000$.

To position the proposed architecture among the state-of-the-art, we compared not only the accuracy but also the training time and number of parameters.
A comparison of CO2 and Power consumption using CodeCarbon~\footnote{https://codecarbon.io/} python library is also presented.

Assessing the utility of LITE in the case of both univariate and multivariate time series classification datasets is essential.
For this reason, we propose two versions: LITE for univariate time series and LITEMV for multivariate time series.
LITEMV differs form LITE only in the first layer, where we use DWSC instead of standard convolutions.
This is not done in the case of the first layer of LITE, when the input is univariate because it would lead to learn only one filter, which is not suitable for the learning task (more details on DWSC are given in Section~\ref{sec:sep_conv}).

Moreover, to address a real world application, we assess the performance of LITE in the case of human rehabilitation exercises, represented as multivariate time series.
For this task, the goal is to analyze the rehabilitation exercise performed by a patient and associate a score representing how good the exercise has been carried out.
We rely on the usage of the Kimore dataset~\cite{capecci2019kimore}, a skeleton based human rehabilitation exercises dataset.
Each sample of this dataset is a
multivariate time series of skeleton poses changing through time.
These skeleton poses are extracted using Kinect-v2 cameras~\cite{lun2015survey_kinect}.
In the original dataset, each skeleton sequence is associated with a continuous label for each sample representing a score between $0$ and $100$ given by an expert to evaluate the performance of the patient.
Since this is a regression problem, we re-orient it to a classification task by classifying whether or not the patient performed the exercise correctly by setting a threshold to divide the scores (threshold score is $50$).
To understand the decision making process of our proposed deep learning model on this task, we also present a study of the interpretability of LITEMV using the Class Activation Map (CAM)~\cite{zhou2016learning,fcn_resnet,ismail2019deep}.

The main contributions of this work
are:
\begin{itemize}
    \item 
    The LITE model is presented as a new architecture for TSC, with only $2.34\%$ of the number of parameters of the Inception model; 
    \item An adaptation of LITE to handle multivariate data (LITEMV);
    \item Extensive experiments showing that LITE achieves state-of-the-art results 
    on the UCR archive
    ;
    \item Extensive experiments showing that LITEMV achieves promising results on some datasets of the UEA archive;
    \item A comparison based on the number of trainable parameters, number of FLOPS, training time, CO2 and Power consumption;
    \item A deeper analysis presented as an ablation study to show the impact of each technique added to boost the proposed model;
    \item The application of LITEMV in a real application of evaluating human rehabilitation exercises with interpretability analysis using CAMs.
\end{itemize}

We note that this manuscript extends our previous work titled``"LITE: Light Inception with boosTing Techniques for Time Series Classification"~\cite{ismail2023lite}. We have expanded our research to include the following enhancements:
\begin{itemize}
    \item A comprehensive analysis on the significant impact of the number of LITE models in the LITETime ensemble;
    \item An extension of the LITE architecture to handle multivariate time series data;
    \item A detailed study on a real world application of human motion rehabilitation, complemented by explainability experiments;
\end{itemize}

In what follows, we present some related work in Section~\ref{sect:related-work}, discuss the details of our proposed architecture in Section~\ref{sec:proposed_method}, present some results compared to other approaches in Section~\ref{sec:results}, and conclude by drawing future work in Section~\ref{sec:conclusion}.

\section{Background and Related Work}\label{sect:related-work}
Time Series Classification (TSC) is widely investigated in the literature.
Some work addressed this problem using machine learning algorithms by comparing similarity metrics between the time series~\cite{bagnall2017great}, decisions based on random forest algorithm~\cite{lucas2019proximity}, \etc~
The problem of most of those algorithms is that they require huge amount of CPU time to perform their calculations, and can not be parallelized on a cluster of GPUs.
However, we acknowledge that libraries such as cuML~\cite{cuml} do exist, which allow for efficient execution of some machine learning workloads on GPUs.
For these reasons, deep learning for TSC is being used in the recent years. 
Even though the accuracy and efficiency of deep learning often exceed those of traditional machine learning algorithms in many contexts, the number of parameters that need to be optimized can be significantly higher.
In what follows, we first define the problem at hand, then we present some works that tackled the TSC problem using machine and deep learning techniques.
Finally, we present some works that addressed the training time problem of deep learning models and the large number of parameters.

\subsection{Definitions}

\begin{itemize}
    \item Let $\textbf{x} = \{x_1,x_2,\ldots,x_L\}$ denote a univariate \emph{time series} of length $L$, consisting of a sequence of data points that are equally spaced in time and exhibit temporal dependencies;
    \item Let $\textbf{X}=\{\textbf{x}_1, \textbf{x}_2, \ldots, \textbf{x}_M\}$ be a multivariate \emph{time series} of $M$ variables, where each variable is itself a univariate time series of length $L$
    \item Let $\mathcal{D}=\{ \textbf{X}_i,y_i\}_{i=0}^{N-1}$ be a \emph{dataset} of $N$ multivariate \emph{time series} $\textbf{X}_i$ with their corresponding labels $y_i$.
\end{itemize}

The goal of this work is to construct an algorithm to learn how to correctly classify each input time series to its corresponding label.

\subsection{Machine Learning for TSC}
The basic approach to solve TSC tasks
is by using the Nearest Neighbour (NN) algorithm.
In order to use this algorithm, a specific similarity metric for time series data should be defined.
In~\cite{bagnall2017great}, the authors used the Dynamic Time Warping algorithm (DTW) to define a similarity metric for the NN algorithm.
However, this algorithm does not have the ability to extract features from the input samples. 
The work in~\cite{zhao2018shapedtw} also used the same algorithm but with an upgraded version of DTW called shapeDTW that aligns a local neighborhood around each point.
The main limitation of the DTW algorithm and its variants is the time complexity, which is a function of the time series length, \ie, $\mathcal{O}(L^2)$.

\subsection{Deep Learning for TSC}
In this section, we present the
work done on TSC using deep learning approaches.
The simplest architecture is the Multi Layer Perceptron (MLP) proposed in~\cite{fcn_resnet} that uses fully connected layers and dropout operations. 
This architecture is limited given the fact that it ignores the temporal dependency in a time series.
The Fully Convolution Network (FCN) was also proposed in~\cite{fcn_resnet} that uses 1D convolution operations.
In this model, the backpropagation algorithm finds the best filters to extract features from the time series, and correctly classifies the samples. In this model, convolutions account for temporal dependencies in time series data, and they are also independent of the input time series \emph{length}.
The authors in~\cite{fcn_resnet} also proposed the ResNet model, which uses the residual connections~\cite{resnet} to solve the vanishing gradient problem. 
A comparative study in~\cite{ismail2019deep} shows that using convolutions, especially ResNet, outperforms other models that use multi-scale transformation or pooling layers with convolutions~\cite{cui2016multi,zheng2016exploiting}.
ResNet and FCN use Batch Normalization and ReLU activation instead of pooling operations after each convolution layer to avoid overfitting.
The state-of-the-art model in deep learning for TSC on the UCR archive~\cite{dau2019ucr} is InceptionTime~\cite{ismail2020inceptiontime}, where the authors adapted the original Inception model on images for time series data.
InceptionTime has the ability to detect multiple patterns of different length given to the multiplexing technique. This technique comes down to learning multiple convolution layers on the same input but with different characteristics.
It is important to note that InceptionTime is an ensemble of five Inception models each trained separately.

While FCN, ResNet have also been evaluated for multivariate data, some additional deep learning models have been proposed to address the task of TSC in the case of multivariate data~\cite{ruiz2021great}.
More recently, the Disjoint-CNN architecture~\cite{foumani2021disjoint}, considered a new methodology for applying convolutions on multivariate time series that separates the temporal and spatial patterns extraction into two steps.
The reference deep learning model for multivariate TSC on the UEA archive~\cite{bagnall2018uea} is ConvTran~\cite{foumani2023improving}, a Transformer based architecture.
ConvTran uses a Disjoint-CNN based encoder followed by self-attention layers~\cite{vaswani2017attention} with novel absolute and relative positional encoders.

Recurrent Neural Networks (RNNs), including their variants LSTM and GRU, have been investigated for Time Series Classification (TSC)~\cite{tanisaro2016time,ismail2019deep}. While these models excel in sequential data processing, they often fail to perform optimally in TSC scenarios due to their design, which does not favor the discriminative identification of complex patterns across time series. This inherent limitation makes them less suited for tasks where precise pattern recognition in time series data is crucial.

Transfer learning and few-shot learning techniques~\cite{fawaz2018transfer,ismail2024finding} can be employed to reduce the training time for models like Inception by leveraging pre-trained networks. While these approaches decrease the duration of training significantly, they do not alter the model's complexity or reduce the number of parameters, which remains a consideration for inference tasks.

\subsection{Reducing Model Complexity}

While deep learning for Time Series Classification (TSC) has proven effective, it faces challenges, such as the extensive number of parameters requiring optimization which can prolong training time. An alternative approach introduced by~\cite{dempster2020rocket} is ROCKET, which utilizes convolution operations but differs significantly from typical deep learning methods like InceptionTime~\cite{ismail2020inceptiontime}. Unlike traditional methods that learn filters through backpropagation, ROCKET employs a strategy of generating a large set ($10,000$) of 1D convolution filters randomly following a Gaussian distribution. These filters vary in length, bias value, padding, and dilation rate, and are not learned but are fixed upon initialization. Each convolution output is then processed to extract two features: the Proportion of Positive Values (PPV) and the maximum value. This results in a substantial feature vector, creating $20,000$ features per series, which are used to train a Ridge classifier. Studies on the UCR archive~\cite{dau2019ucr} indicate no statistical significant difference in accuracy between InceptionTime and ROCKET, yet ROCKET offers a substantial advantage in terms of both training and inference speed.

Some adaptations of ROCKET were proposed in order to boost its performance even more such as MiniROCKET~\cite{dempster2021minirocket} and MultiROCKET~\cite{tan2022multirocket}.
Knowledge distillation~\cite{hinton2015distilling} was also approached for the TSC model called FCN~\cite{ay2022study}. In this study, the authors proposed a smaller variant of FCN with a lower number of convolution layers and filters to learn.

Furthermore, the work in~\cite{ismail-fawaz2022hccf} proposed to hand-craft some custom convolution filters instead of randomly generate them. Those hand-crafted filters are constructed in a way to get activated on increasing and decreasing intervals as well as peaks in the time series.
By using these filters, the authors were able to construct a Hybrid FCN (H-FCN). Results on the UCR archive shown that H-FCN is statistically significantly better than FCN and is competitive with InceptionTime.
The H-FCN model uses the custom filters in parallel to the learned filters in the first layer.

Some work to optimize the complexity of large models were proposed in Computer Vision as well.
For instance, the authors in~\cite{howard2017mobilenets} proposed the usage of DWSC instead of standard ones.

The MobileNet architecture as proposed in~\cite{howard2017mobilenets} proven to be very competitive with state-of-the-art models on ImageNet~\cite{deng2009imagenet} with way less complexity.

In Natural Language Processing, some works proposed the usage of Small Language Models (SLM) as a one-shot learning approach~\cite{schick2020s}. The authors showed that with way less parameters than GPT-3~\cite{brown2020language}, their model can have no significant difference in performance.

\section{Proposed approach}\label{sec:proposed_method}

\subsection{Convolutions for TSC}
Multiple Convolution Neural Networks (CNNs) were proposed for the task of TSC and they all proved how they outperform other methods.
The Fully Convolution Network (FCN) is a simple three layered network, where each layer is composed of 1D convolutions followed by a batch normalization and a ReLU activation function.
As FCN is only composed of simple 1D convolution layers, its performance generally lags behind more advanced architectures using residual connections (ResNet) or Inception modules (InceptionTime).
In this paper, we present an adaptation of the FCN architecture that only has $2.34\%$ of the number of parameters of the Inception model. 
Given this significant drop in the number of parameters, boosting techniques are used in order to preserve the performance of Inception.

First, we discuss about two ways of applying convolutions with less number of parameters, while preserving the performance. The first approach uses standard convolutions with BottleNecks (BN), and the second uses DWSC.

\subsubsection{Standard Convolutions with BottleNecks}
Many approaches that use CNN based architectures suffer from the problem of high number of parameters such as ResNet~\cite{fcn_resnet}.
For this reason, the authors of InceptionTime~\cite{ismail2020inceptiontime} proposed to use a BottleNeck operation in order to reduce the number of parameters.
This BottleNeck operation is made of 1D convolutions with a unit kernel size. 

The following example shows the impact of this operation on reducing the number of parameters.
Suppose at a depth $d$ in the network, the input number of channels is $C_{in}$. The following convolution layer of depth $d+1$ projects the input into a new space with a number of channels $C_{out}$ using a kernel of size $k$. On the one hand, without a BottleNeck operation, the number of learned parameters is $C_{in}*C_{out}*k$. On the other hand, with a BottleNeck operation that uses $C_{bn}$ filters of size $1$, the number of learned parameters is $C_{in}*C_{bn}*1+C_{bn}*C_{out}*k$.
This operation reduces the number of parameters if and only if the following inequality is true:
\begin{equation}
    C_{in}*C_{bn} + C_{bn}*C_{out}*k < C_{in}*C_{out}*k ,
\end{equation}

\noindent
which indicates that the condition on the BottleNeck operation is:
\begin{equation}
    C_{bn} < \frac{C_{in}*C_{out}*k}{C_{in} + C_{out}*k}.
\end{equation}

The goal of the BottleNeck operation is to learn the same number of filters in the output channels ($C_{out}$), while reducing, at the same time,
the intermediate learned filters between the input and output channels ($C_{in}*C_{out}$).

\subsubsection{DepthWise Separable Convolution DWSC}\label{sec:sep_conv}
\begin{figure}[!ht]
    \centering
    \includegraphics[width=0.7\linewidth]{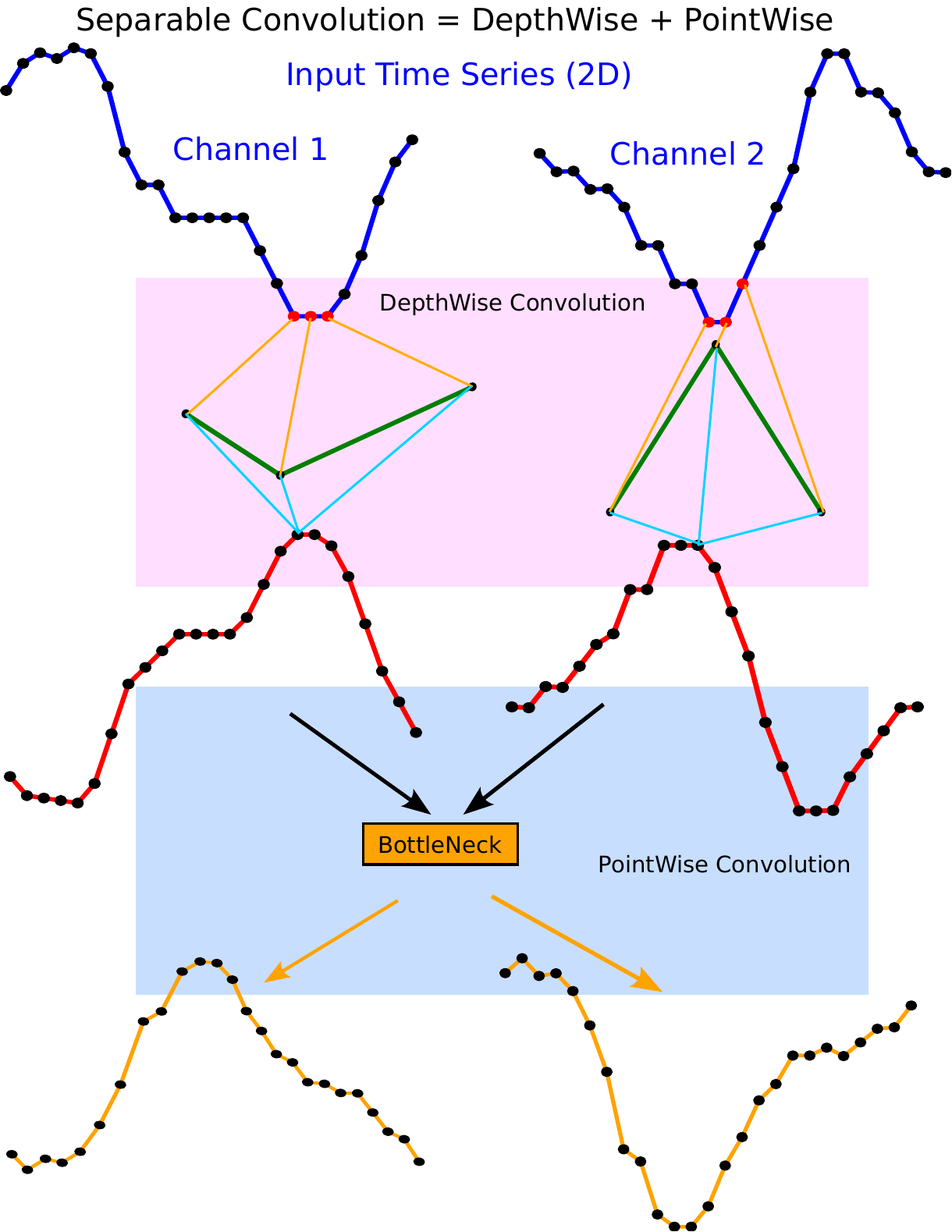}
    \caption{DWSC for time series represented in its two phases: (1) DepthWise convolution (\colorbox{purpleF2}{purple} block), and (2) PointWise convolution (\colorbox{blueL}{blue} block).}
    \label{fig:sep-conv}
\end{figure}

DWSC can be divided into two phases: DepthWise convolution (\emph{Phase 1}), and PointWise convolution (\emph{Phase 2}).
A visual representation of the DWSC operation is presented in Figure~\ref{fig:sep-conv}.

In standard convolutions, if the input sample of length $L$ has $C_{in}$ channels and the desired output is a space with $C_{out}$ channels using a kernel of size $k$, then the number of learned parameters is $C_{in}*C_{out}*k$. The number of multiplications is $L*C{out}*C_{in}*k$.
This assumes, as with other parts of this section, that the convolution layer uses ``same'' padding, thereby preserving the length from the input series to the output.

\paragraph{DepthWise convolution} In this phase, if the convolution is done using a kernel of size $k$, then the number of learned filters is $C_{in}$, and the output number of channels will be $C_{in}$ (\emph{such as the input}). In other words, for each dimension of the input time series, one filter is learned.

\paragraph{PointWise convolution} This phase projects the output of the DepthWise convolution into a space with a desired number of channels $C_{out}$. This is done by applying a standard convolution with $C_{out}$ filters of kernel size 1 (\emph{a BottleNeck}).

Equation~\ref{equ:dwsc-final} below presents the overall formulation of a DWSC layer applied on an input multivariate time series $\textbf{X}=\{\textbf{x}_1,\textbf{x}_2,\ldots,\textbf{x}_M\}$ with $M$ variables of length $L$, with $C_{out}$ as a target output dimension using a kernel size of length $k$.
\begin{equation}\label{equ:dwsc-final}
    \textbf{O} = concat(\{\sum_{m=1}^{M}concat(\{\textbf{x}_m*\textbf{w}_m\}_{m=1}^{M})^m.w_{m,j}\}_{j=1}^{C_{out}})
\end{equation}
\noindent where $\textbf{w}_m$ is the $m^{th}$ convolution filter applied in phase 1 (DepthWise convolution), $concat(.)$ is the concatenation operation over the spatial dimension, $*$ is the one dimensional convolution operation and $w_{m,j}$ is one trainable scalar of phase 2 (PointWise convolution) where $m~\in~[1,M]$ and $i~\in~[1,C_{out}]$, making the output of the convolution $\textbf{O}$ a multivariate time series of $C_{out}$ dimensions of length $L$ assuming a padding ``same''.

Hence, by combining these two phases, the number of learned parameters in a DWSC becomes $C_{in}*k + C_{in}*C_{out}$.

The following calculation finds the condition that the DWSC have less parameters to learn compared to the standard one:
\begin{equation}
    \begin{split}
        \underbrace{C_{in}*C_{out}*k}_\text{standard Conv} &> \underbrace{C_{in}*k + C_{in}*C_{out}}_\text{separable Conv}\\
        C_{in}*C_{out}*k &> C_{in}*(k+C_{out})\\
        C_{out}*k &> k + C_{out}\\
        k*(C_{out} - 1) &> C_{out}\\
        k &> \dfrac{C_{out}}{C_{out} - 1} \xrightarrow{C_{out}\to \infty} \boxed{k > 1}.
    \end{split}
\end{equation}

This means that if the number of desired output channels is high enough (if $C_{out} >= 3$ the previous equation holds), then DWSC have less parameters to learn compared to the standard convolutions.

The number of multiplications performed in the DWSC is $C_{in}*L*k + C{in}*C_{out}*L*1$. The following calculation finds the second condition for when DWSC have less multiplications to perform compared to standard convolutions:
\begin{equation}
    \begin{split}
        \underbrace{C_{in}*C_{out}*L*k}_\text{standard convs} &> \underbrace{C_{in}*L*k + C_{in}*C_{out}*L*1}_\text{separable convs}\\
        C_{in}*C_{out}*L*k &> C_{in}*L*(k + C_{out})\\
        C_{out}*k &> k + C_{out}\\
        k(C_{out} - 1) &> C_{out}\\
        k &> \dfrac{C_{out}}{C_{out}-1} \xrightarrow{C_{out}\to \infty} \boxed{k > 1}.
    \end{split}
\end{equation}

This concludes that DWSC have less parameters to learn with less number of multiplications to perform compared to standard convolutions.
In this work, we present a comparison between the usage of DWSC or standard ones + BottleNecks.
Our results demonstrate that with the techniques added to boost DWSC, we can maintain performance, while significantly reducing the number of parameters to optimize.

After defining two techniques to use convolutions in a more optimized way concerning number of parameters and multiplications, some other techniques should be defined as well.
These techniques aim to minimize the impact of parameters reduction in convolutions operations explained above.

\subsection{Boosting Techniques}
The following techniques are borrowed from the literature.

\begin{figure}[!ht]
    \centering
    \includegraphics[width=0.8\linewidth]{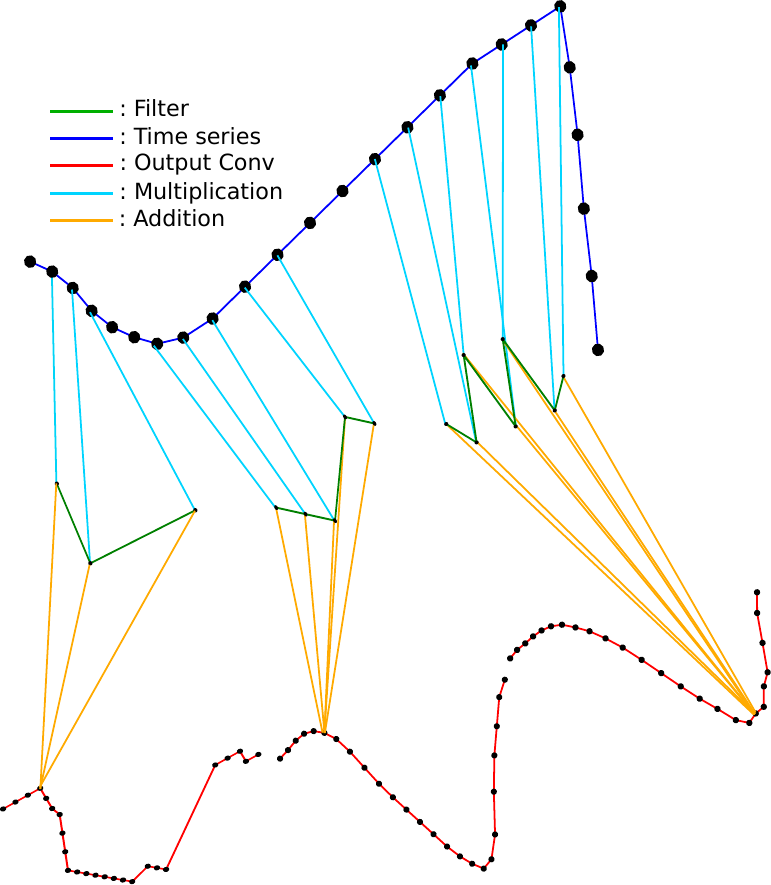}
    \caption{Multiplexing one dimensional convolution on the input time series (in blue) using filters (in \colorbox{greenF2}{green}) with three different kernel sizes, respectively, 3, 5, and 7.
    The output of the convolutions (in \colorbox{redF2}{red}) is different for each filter.}
    \label{fig:multiplexing}
\end{figure}

\paragraph{Multiplexing}
Multiplex convolution was proposed in the architecture of Inception~\cite{ismail2020inceptiontime}. Its main idea is to learn in parallel different convolution layers of different kernel size.
A multiplexing example is shown in Figure~\ref{fig:multiplexing}.

\begin{figure}[!ht]
    \centering
    \includegraphics[width=0.8\linewidth]{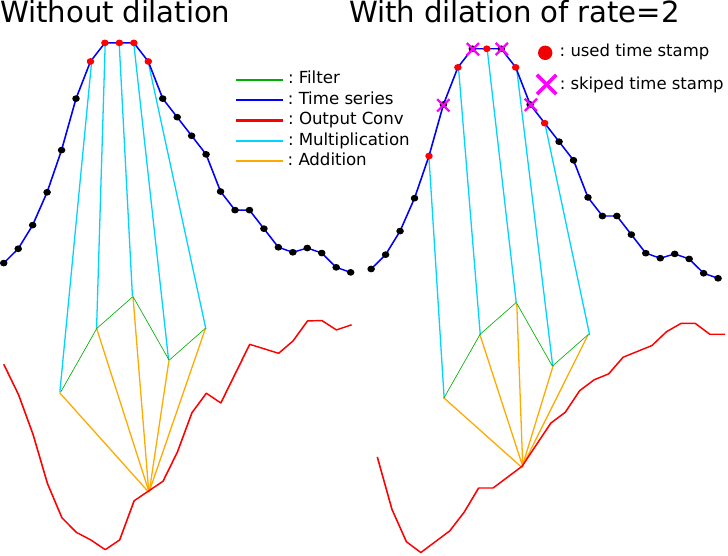}
    \caption{One dimensional convolution performed w/o dilation on the left (\emph{rate}=1), and with dilation on the right (\emph{rate}=2).}
    \label{fig:dilation}
\end{figure}

\paragraph{Dilation}
Dilated convolutions were not very explored for deep supervised learning on TSC when evaluating on the UCR archive~\cite{dau2019ucr} but they were used in self-supervised models showing to be very effective~\cite{mixingUp}.
Additionally, outside the score of the UCR archive~\cite{dau2019ucr}, some research work have addressed the task of deep supervised learning on TSC using dilated convolutions~\cite{risso2022lightweight,zanghieri2019robust,bai2018empirical}.
Dilation in convolutions filters defines the number of empty cells in the kernel. Suppose a kernel of length $3$ has the following parameters $k = [k_0,k_1,k_2]$, a dilation of rate 2 will produce the following kernel $k = [k_0, skip, k_1, skip, k_2]$.
The $skip$ parameter indicates that the convolution layer will not use the values of the input aligned with this index of the kernel.
A visualization of the dilation effect on convolution can be seen in Figure~\ref{fig:dilation}.
Dilation will help increasing the receptive field of a model without having to add deeper layers because the dilated kernel will find the deeper combinations in the same layer.

\paragraph{Custom Filters}
Custom filters were proposed in~\cite{ismail-fawaz2022hccf}. The authors hand-crafted some kernels in order to detect specific patterns in the input time series. These filters were then added to Inception and results on the UCR archive have shown that such filters can help with the generalization and boost the performance. This is due to the fact that these filters are generic and fixed (not learned).
This allows the model to focus on learning new patterns harder to detect.

\subsection{Proposed architecture}
\subsubsection{Light Inception with boosTing tEchniques (LITE)}\label{sec:LITE}
\begin{figure*}
    \centering
    \includegraphics[width=0.7\textwidth]{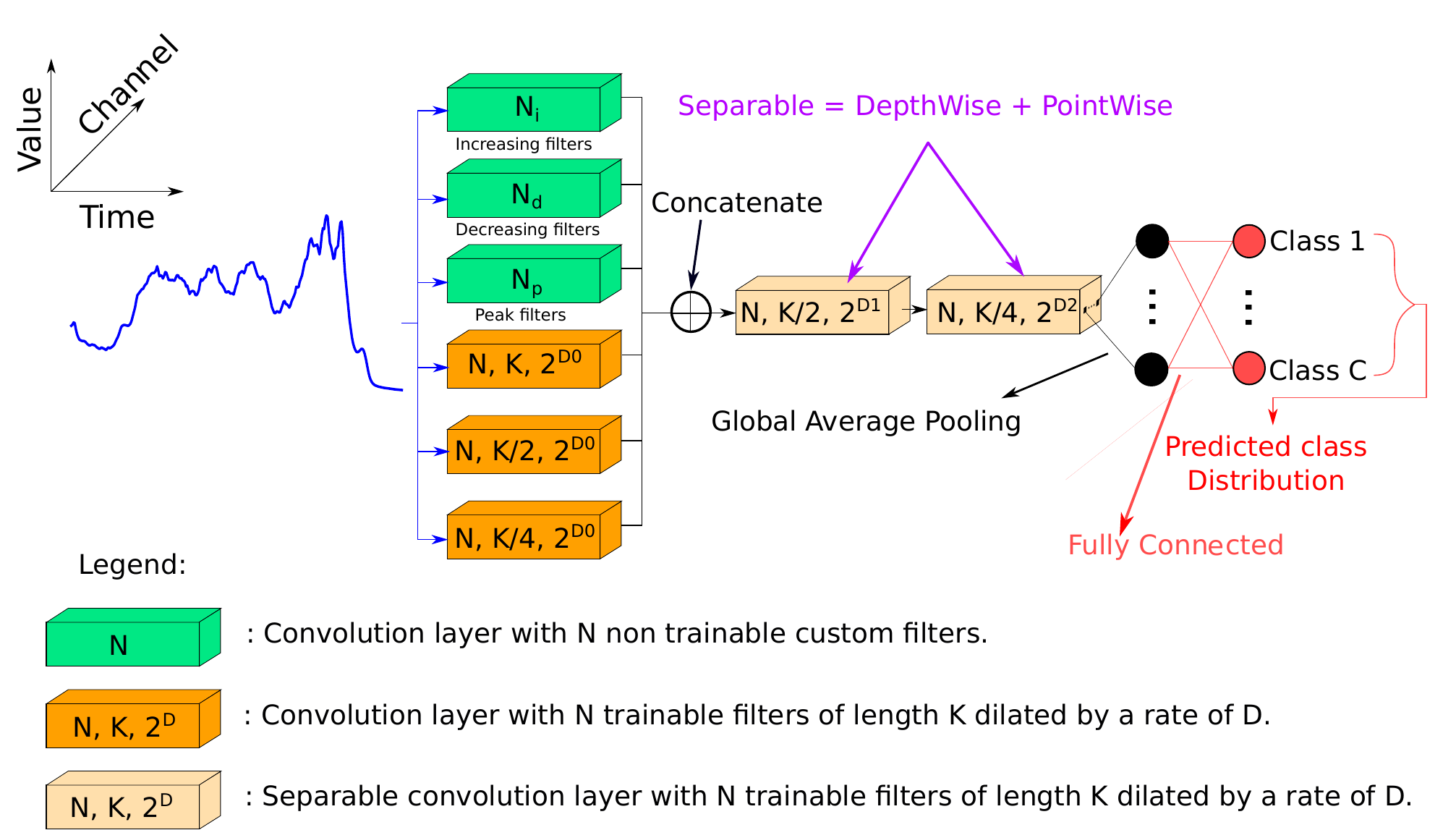}
    \caption{The proposed LITE architecture that uses multiplexing convolutions in the first layer (three convolution blocks in \colorbox{orangeF2}{orange}) with custom filters (in \colorbox{greenF2}{green}). The second and third layer are composed of DWSC (in \colorbox{beigeF2}{beige}). The last layer is followed by a Global Average Pooling (GAP) on the time axis and finished by a classification Fully Connected layer to approximate the class distribution.}
    \label{fig:LITE}
\end{figure*}

In our proposed architecture, we reduce the number of parameters, while preserving performance. This is obtained by using DWSC and the previously explained boosting techniques.
First, custom filters are used in the first layer parallel to the first layer. Second, in this first layer, multiplexing convolution is used in order to detect different patterns of different characteristics (three parallel convolution layers). Third, the second and third layer present the usage of dilation in their kernels.
It is important to notice that for the first layer, standard convolutions are used instead of DWSC. This is due to the fact that the input time series is univariate and DWSC will learn only one filter.
A summary of the architecture is given in Figure~\ref{fig:LITE}.

\subsubsection{LITEMV: Adapting LITE for Multivariate Time Series Data}
The first layer of the LITE architecture uses the standard convolution in the case of univariate data.
This is due to the fact that using DepthWise Separable Convolutions in the first layer means the model would only learn one filter and then re-scale the output with one learned value.
For this reason we kept the usage of Standard Convolutions in the first layer.
However, in the case of multivariate data, the model will learn a filter per channel and then learn how to combine in the PointWise Convolution step.
A second information that is important to note is that the custom filters in the first layer of LITE are basically Standard Convolutions.
This means in the case of multivariate input, the custom filter will be applied to each channel independently and the output are summed.
This is done for each of the used custom filters.
This can be an issue in the case of multivariate input given we suppose an equal importance per custom filter for all channels when summing the outputs.
In the multivariate version of LITE, we instead concatenate the outputs of all custom filters operations on all channels.
We refer in the rest of this work to the multivariate version of LITE as LITEMV.

\subsubsection{Ensemble}
Ensemble learning is a technique of combining the prediction of multiple models in order to reduce the variance, and it has been shown to be very effective in the literature~\cite{fawaz2019deep,ismail2020inceptiontime,middlehurst2021hive}. Applying an ensemble of multiple classification models is equivalent to find the average predicted distribution of all the models.
This average distribution is finally used for choosing the predicted class.
This motivated us to build an ensemble of multiple LITE models to form LITETime 
and LITEMVTime
(by adding the suffix \emph{Time} following~\cite{ismail2020inceptiontime}). Moreover, the usage of an ensemble in the case of LITE is also motivated by its small architecture and the fact that it can boost less complex architectures even more.

\section{Experimental Evaluation}\label{sec:results}

\subsection{Datasets}
The UCR archive~\cite{dau2019ucr} is the largest directory for the TSC problem. It is publicly available~\cite{dau2019ucr}.
The archive contains 128 datasets of univariate TSC tasks.
Some tasks involves Electrocardiography (ECG) time series data and some are observations of Sensors, \etc~
To evaluate the performance of the proposed architecture on more datasets, we detail in Section~\ref{sec:results_uea} some experiments on multivariate time series data.
The usage of multivariate data is a bit different given the uniqueness of this data of having multiple variables changing through time.
We rely on the UEA Multivariate Time Series Classification (MTSC) archive~\cite{bagnall2018uea}.
This archive consists of 30 different datasets ranging from medical data recording ECG signals to speech recognition.
The number of channels in this archive range from $2$ to $1345$.

Each dataset is split into a training and a testing set. The labels are available for all the samples. In order to train the model on a normalized dataset, we apply $z$-normalization over all the samples independently. This normalization technique reduces the time series samples to a zero mean and unit variance sequence.

\subsection{Implementation Details}
Our results were obtained on the UCR archive using a GTX 1080 GPU with 8GB of VRAM.
In the experiments, we accounted for the training time, inference time (testing time), CO2 and Power consumption.
The model used for testing is the best model during training, chosen by monitoring the training loss. The Adam optimizer was used with a Reduce on Plateau learning rate decay method by monitoring the training loss. For the Adam optimizer we used the default set up of the Tensorflow Python model. The model is trained with a batch size of $64$ for $1500$ epochs.

For the hyper-parameters of the LITE architecture presented in Figure~\ref{fig:LITE}, the following setup was used:
$N_i=6$, variations in kernel sizes $=[2,4,8,16,32,64]$; $N_d=6$, variations in kernel sizes $=[2,4,8,16,32,64]$; $N_p=5$, variations in kernel sizes $=[6,12,24,48,96]$; $N=32$; $K=40$; $D0=1$ in order to start with no dilation and increase with depth ($D1=2$ and $D2=4$).
The same hyper-parameters were used in the case of LITEMV.
The source code is publicly available at~\texttt{https://github.com/MSD-IRIMAS/LITE}.

\begin{table*}
\centering
\caption{Comparison between the proposed methods with FCN, ResNet and Inception without ensemble.}
\label{tab:PTCE}
\begin{tabular}{|c|c|c|c|c|c|c|}
\hline
Models &
  \begin{tabular}[c]{@{}c@{}}Number of\\ parameters\end{tabular} &
  FLOPS &
  Training Time &
  Testing Time &
  $CO2_{eq}$ &
  Energy \\ \hline
Inception &
  420,192 &
  424,414 &
  \begin{tabular}[c]{@{}c@{}}145,267 seconds\\ 1.68 days\end{tabular} &
  \begin{tabular}[c]{@{}c@{}}81 seconds\\ 0.0009 days\end{tabular} &
  0.2928 g &
  0.6886 kWh \\ \hline
  ResNet &
  504,000 &
  507,818 &
  \begin{tabular}[c]{@{}c@{}} 165,089 seconds\\  1.91 days\end{tabular} &
  \begin{tabular}[c]{@{}c@{}} 62 seconds\\  0.0007 days\end{tabular} &
   0.3101 g &
   0.7303 kWh \\ \hline
FCN &
  264,704 &
  266,850 &
  \begin{tabular}[c]{@{}c@{}}149,821 seconds\\ 1.73 days\end{tabular} &
  \begin{tabular}[c]{@{}c@{}}27 seconds\\ 0.00031 days\end{tabular} &
  0.2623 g &
  0.6176 kWh \\ \hline
\textbf{LITE} &
  \textbf{9,814} &
  \textbf{10,632} &
  \begin{tabular}[c]{@{}c@{}}\textbf{53,567 seconds}\\ \textbf{0.62 days}\end{tabular} &
  \begin{tabular}[c]{@{}c@{}}\textbf{44 seconds}\\ \textbf{0.0005 days}\end{tabular} &
  \textbf{0.1048 g} &
  \textbf{0.2468 kWh} \\ \hline
\end{tabular}
\end{table*}

\subsection{Results and Discussion}

Below, we provide the comprehensive set of experimental results acquired during the course of this study.
First, we present an efficiency comparison between LITE and other architectures in the literature.
This study is done over the case of univariate data only because in the case of multivariate data the study will depend on each dataset and its number of channels.
Second, we present an ablation study to make sure that LITE utilizes all of its boosting techniques during training.
Third, we present the performance using the accuracy metric on both the UCR and UEA archives.
Finally, we evaluate the usage of LITEMVTime in a real world application for human motion rehabilitation on the Kimore dataset.

\subsubsection{Number of Parameters, FLOPS Training Time, Testing time, CO2 and Power Consumption}

Table~\ref{tab:PTCE} summarizes the number of parameters, the number of FLoating-point Operations Per Second (FLOPS), training time, inference time, CO2 and Power consumption over the 128 datasets of the UCR archive.
We show the number of trainable parameters of the architecture without the last classification Fully Connected layer because it depends on each dataset (number of classes). The rest of the information is summed over the 128 datasets of the UCR archive and averaged over five different runs.

First, the table shows that the smallest model in terms of number of parameters is the LITE with $9,814$ parameters.
This is mainly due to the usage of DWSC instead of standard ones.
Compared to FCN ResNet and Inception, LITE has only $3.7\%$ $1.95\%$ and $2.34\%$ of their number of parameters, respectively.
Second, the fastest model in the training phase is LITE, with a training time of $0.62$ days.
LITE is $2.79$, $3.08$ and $2.71$ time faster than FCN, ResNet and Inception, respectively. 
Third, LITE is the model that consumes the smallest amount of CO2 and Power, $0.1048$ grams and $0.2468$ Watts, respectively.
Compared to the other approaches, LITE presents the fastest and most ecologic model for TSC compared to FCN, ResNet and Inception.
We believe, given the factors explained above, that LITE can be used for the deployment of deep learning for TSC in small machine such as mobile phones.
As shown in Table~\ref{tab:PTCE}, the LITE model not only has the smallest number of parameters among the evaluated models but also has the second fastest inference time, making it well-suited for environments where model size and speed are crucial.

Although DTW is a relevant technique in time series analysis, it was not included in our comparative efficiency analysis due to its inability to be parallelized on GPUs, a feature that our proposed deep learning approaches exploit to achieve scalability and speed.
Additionally, ROCKET was excluded from our efficiency comparative analysis because it is primarily implemented for multi-threaded CPU use, not GPU acceleration, which differs from the hardware focus of our study.

\subsubsection{Accuracy Performance on the UCR Archive}\label{sec:acc-perf}
\begin{figure*}
    \centering
    \includegraphics[width=0.7\textwidth]{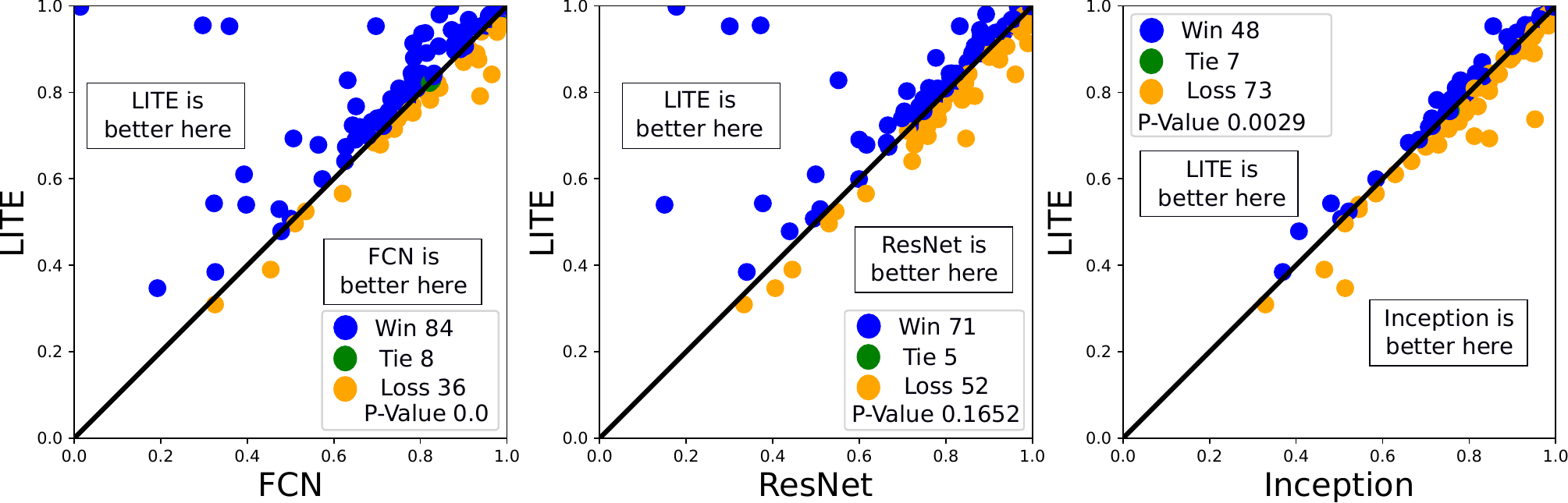}
    \caption{One-vs-One comparison between LITE with three different models: FCN, ResNet and Inception over the 128 datasets of the UCR archive.}
    \label{fig:res_LITE}
\end{figure*}
\begin{figure}[!ht]
    \centering
    \includegraphics[width=\linewidth]{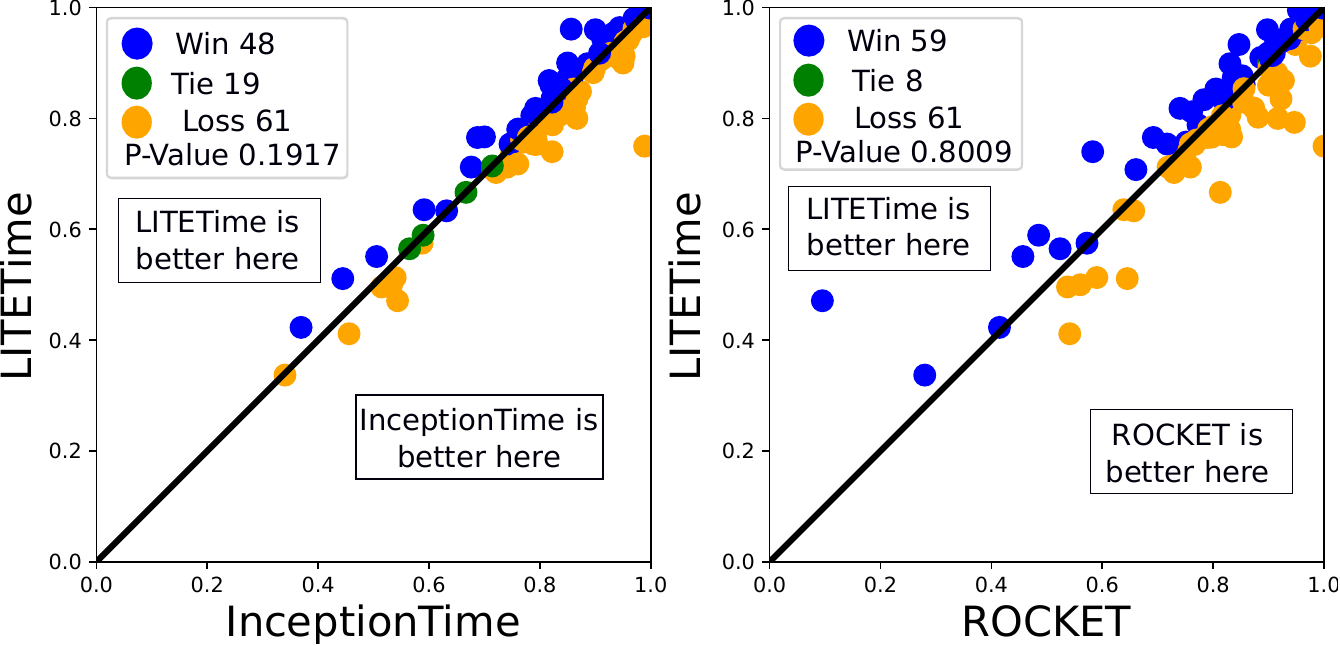}
    \caption{One-vs-One comparison between LITETime with two different models: InceptionTime and ROCKET over the 128 datasets of the UCR archive.}
    \label{fig:res_LITEtime}
\end{figure}

In what follows, a one-vs-one comparison is 
presented between the models in order to show that LITE can preserve the performance of the more complex architectures.
This one-vs-one comparison comes down to a Win/Tie/Loss count on the 128 datasets of the UCR archive between two classifiers.
This comparison is visualized in Figures~\ref{fig:res_LITE} and~\ref{fig:res_LITEtime}. Each point in these plots represents one dataset of the UCR archive. The $x$-axis contains the accuracy value on the test set using classifier-$x$ and the $y$-axis the ones using classifier-$y$.

In order to evaluate the significance of the Win/Tie/Loss comparison, a statistical Wilcoxon Signed Rank Test~\cite{wilcoxon1992individual} is used. This test will return a statistical value, the P-Value, representing how significant the difference is between the two classifiers. If the $P$-Value is low, this would mean that the difference in performance between the two classifiers is statistically significant.
If the last condition is not true, it means that there are not enough examples (datasets) to find a statistical significant difference between the classifiers. This Wilcoxon test needs a P-Value threshold, usually in the literature a $0.05$ (or $5\%$) threshold is used.

\begin{table*}
\centering
\caption{Datasets where Inception beats LITE by more than $10\%$ of accuracy.}
\label{tab:inception-better}
\begin{tabular}{|c|ccc|}
\hline
\begin{tabular}[c]{@{}c@{}}Dataset Name\\ where Inception\\ is better than LITE\end{tabular} &
  \multicolumn{1}{c|}{\begin{tabular}[c]{@{}c@{}}Difference in\\ Accuracy (\%)\end{tabular}} &
  \multicolumn{1}{c|}{\begin{tabular}[c]{@{}c@{}}Series Length\end{tabular}} &
  \begin{tabular}[c]{@{}c@{}}Number of\\ Samples\end{tabular} \\ 
  \hline
EthanolLevel      & 11.32 & 1751 & 504 \\
OliveOil          & 15.34 & 570 & 30 \\
PigAirwayPressure & 16.64 & 2000  & 104  \\
ShapeletSim       & 21.44 & 500  & 20  \\ 
\hline
\end{tabular}
\end{table*}

On the one hand, the results presented in Figure~\ref{fig:res_LITE} show that LITE beats FCN significantly (low $P$-Value), and is statistically not significant than ResNet (high $P$-Value).
The results compared to ResNet are impressive given the small complexity of LITE (almost $1.95\%$ of ResNet's number of parameters). On the other hand, the comparison shows that LITE still is not significantly close to Inception.
To study more the reason why LITE performs not as good as Inception with a large margin (more than $10\%$), we presented some characteristics of those datasets in Table~\ref{tab:inception-better}.
This table shows that some of the datasets have either long time series or small training set.
Firstly, this indicates that Inception is
better than LITE on long time series given its large receptive field (deeper architecture).
Secondly, this demonstrates that the Inception model generally exhibits stronger generalization capabilities, particularly when the dataset has a smaller training set.

Given that Inception still beats LITE, an ensemble comparison shows the real performance of the proposed architecture.
This is due to the fact that LITE has way less parameters ($2.34\%$ of Inception's number of parameters), which can make it sensitive to a higher variance when training with different initialization.
Applying an ensemble removes this variance as explained before.

The comparison between LITETime with InceptionTime and ROCKET is presented in Figure~\ref{fig:res_LITEtime}. This comparison shows that, given the 128 datasets of the UCR archive, there are not enough datasets to find a statistical significance in the difference of performance with InceptionTime and ROCKET. We included ROCKET in the ensemble comparison with LITETime because the motivation in ROCKET is to replace the ensemble technique by using random filters instead of learning them starting with different initialization.

Those last results suggest that in order to get a good performance on the UCR archive, a large complex architecture with a high number of parameters is not always needed.

\begin{figure}[!ht]
    \centering
    \includegraphics[width=\linewidth]{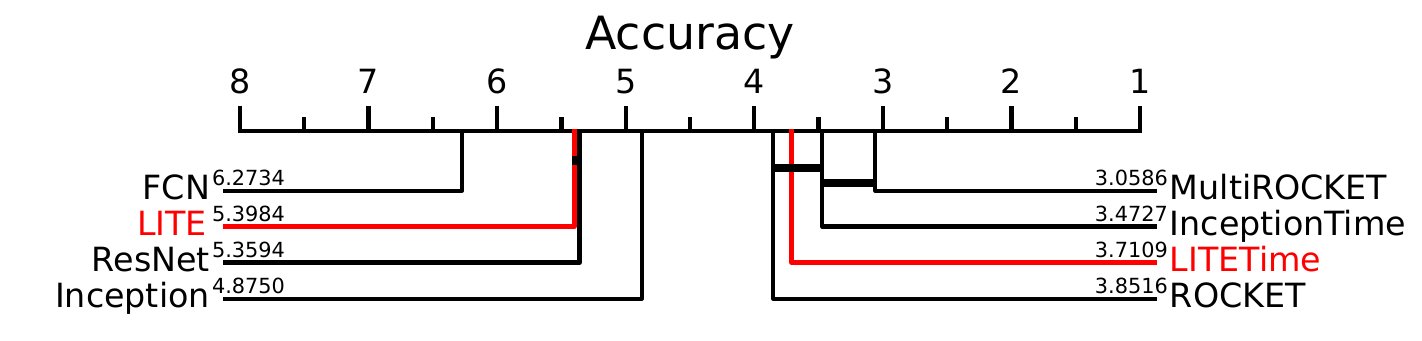}
    \caption{A Critical Difference Diagram (CD Diagram) showing the average rank of each classifier over the 128 datasets of the UCR archive with the significance in difference of performance.}
    \label{fig:res_cdd}
\end{figure}

For a multi-classifier comparison, the average rank of each model is shown in a Critical Difference Diagram~\cite{demvsar2006statistical} (CD Diagram) based on the ranking classifiers given the average rank over multiple datasets, the two tailed Wilcoxon Signed-Rank Test with the Holm multiple test correction~\cite{benavoli2016should}.
To generate the CD Diagram, we used the publicly available code~\texttt{https://github.com/hfawaz/cd-diagram}.
This diagram also presents connections between classifiers, when the difference in performance is not statistically different following the Wilcoxon Signed Rank Test. A CD Diagram is presented in Figure~\ref{fig:res_cdd} and shows that LITETime comes $3rd$ on the average rank between ROCKET and InceptionTime.
The diagram also shows that FCN performs statistically significantly worse than LITE on the average rank.
Furthermore, on the UCR archive, no statistical significance can be observed between ResNet and LITE.
This last comparison shows the real impact of this work where LITE has almost $1.95\%$ (Table~\ref{tab:PTCE}) of ResNet's number of parameters.
These conducted results show that on a large amount of cases, there is no need for a complex model with high number of parameters to achieve good performance.

\begin{figure*}
    \centering
    \includegraphics[width=\textwidth]{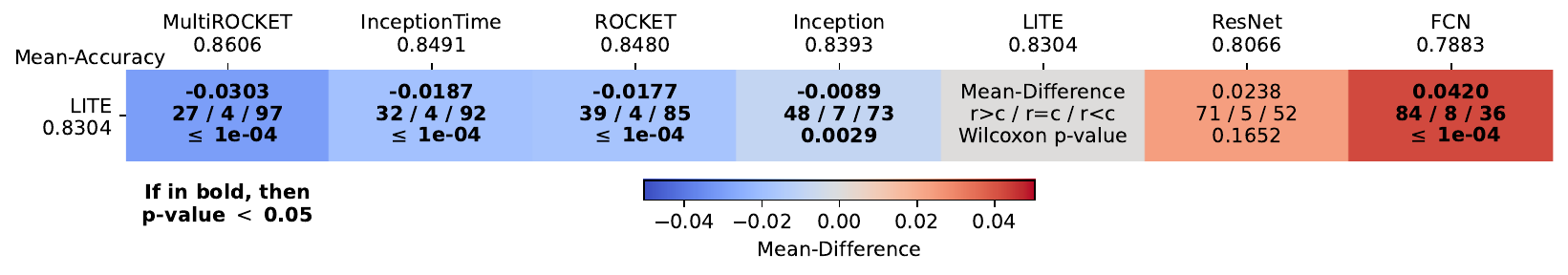}
    \caption{The Multi-Comparison Matrix applied to show the performance of LITE compared to other approaches.}
    \label{fig:mcm-lite}
\end{figure*}

\begin{figure*}
    \centering
    \includegraphics[width=\textwidth]{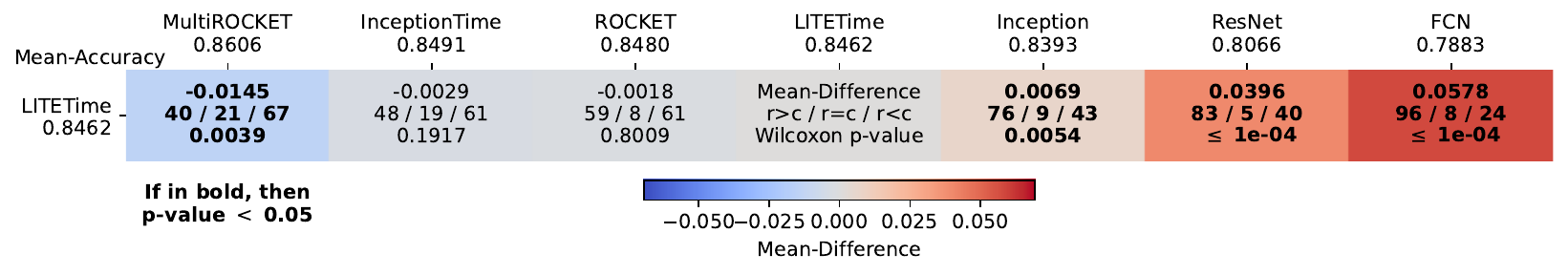}
    \caption{The Multi-Comparison Matrix applied to show the performance of LITETime compared to other approaches.}
    \label{fig:mcm-litetime}
\end{figure*}

\begin{figure*}
    \centering
    \includegraphics[width=\textwidth]{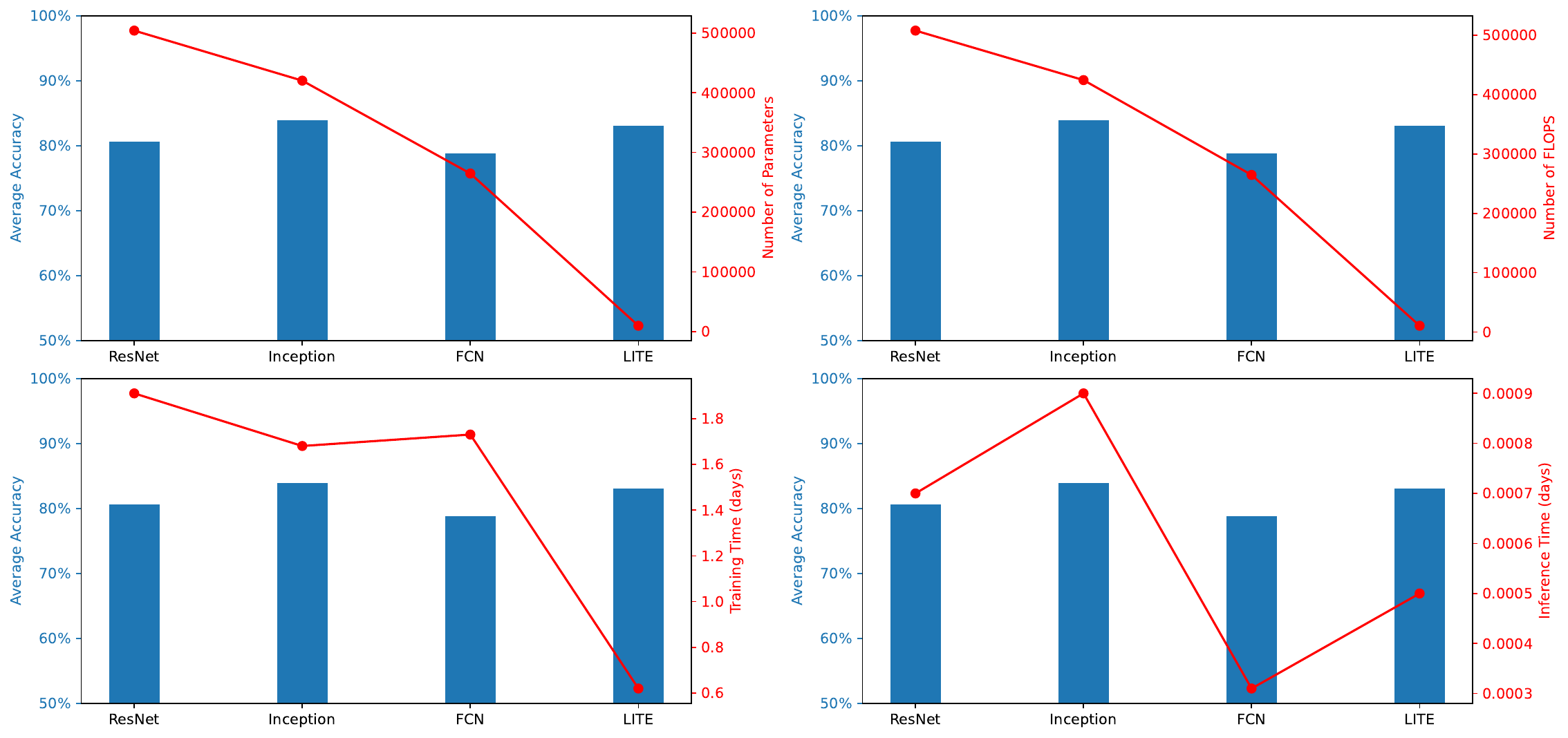}
    \caption{A collection of four Pareto plots showcasing the average accuracy of the four models ResNet~\cite{fcn_resnet} Inception~\cite{ismail2020inceptiontime}, FCN~\cite{fcn_resnet} and ours LITE, over the UCR archive datasets~\cite{dau2019ucr} in function of the number of parameters, number of flops, training time (in days) and inference time (in days).}
    \label{fig:ptce}
\end{figure*}

Furthermore, a new Multi-Comparison Matrix (MCM) evaluation tool was proposed that is stable to the variation of the addition and removal of classifiers~\cite{ismail2023approach}.
The MCM is presented in Figures~\ref{fig:mcm-lite} and~\ref{fig:mcm-litetime} to compare LITE and LITETime, respectively, to other approaches in the literature.
The MCM uses the average accuracy on the UCR as an ordering metric instead of the average rank.
As presented in the MCMs, LITE and LITETime perform better than FCN and ResNet on the average accuracy and is closer to the performance of InceptionTime, which is not significantly different than LITETime (high p-value).
MCM has an advantage over the usage of the CD Diagram of being stable with the addition and removal of classifiers, given that the average performance would not change in this scenario unlike the average rank.
Another advantage is not using a multiple test correction for the $P$-Value significance test.

As depicted in the MCM in Figure~\ref{fig:mcm-litetime}, LITETime outperforms Inception with statistical significance in terms of accuracy. Regarding latency, although a single LITE model operates at approximately half the inference time of Inception, the combined inference time of the LITETime ensemble, which includes five LITE models, is about 2.5 times that of Inception. However, it is crucial to note that the ensemble's overall complexity in terms of the number of parameters is significantly reduced—eight times smaller than Inception. This substantial reduction in model complexity, coupled with the potential for parallelizing these models, suggests that improvements in parallel processing could effectively diminish the apparent latency disadvantage, offering a practical solution for scenarios where model efficiency and reduced hardware demand are critical.

In addition to Table~\ref{tab:PTCE}, Figure~\ref{fig:ptce} presents similar metrics, including a collection of Pareto plots comparing the average performance of LITE and three comoetitotrs in function of number of parameters, number of FLOPS, training time and infernece time.
The plot illustrates that LITE performs as good as state-of-the-art models, such as Inception with a difference in average accuracy of only $0.89\%$, while being $42$ times smaller than Inception, having $40$ times less FLOPS, and being $2.7$ and $1.8$ times faster in terms of training and inference time respectively.

\subsection{Ablation Study - LITE}

\subsubsection{Impact of Additional Techniques}
\begin{figure*}
    \centering
    \includegraphics[width=0.7\textwidth]{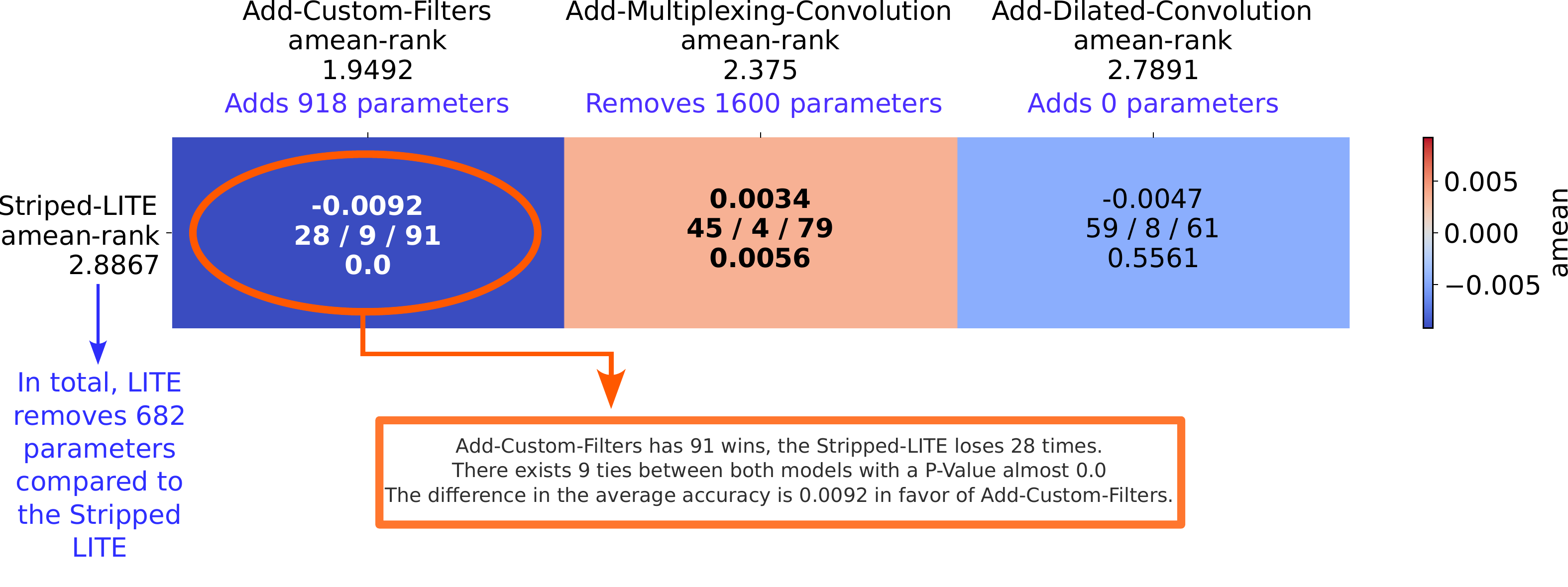}
    \caption{The Heat Map shows the one-vs-one comparison between the Striped-LITE and the three variants: (1) Add-Custom-Filters, (2) Add-Multiplexing-Convolution and (3) Add-Dilated-Convolution. The colors of the Heat Map follow the value of the first line in each cell. This value is the difference between the value of the first line (average accuracy when winning/losing). The second line represents the Win/Tie/Loss count between the models in question (wins for the column model). The last line is the statistical $P$-Value between the two classifier using the Wilcoxon Signed Rank Test.}
    \label{fig:ablation_LITE_results}
\end{figure*}

The proposed LITE architecture, uses multiple techniques in order to improve the performance.
In order to show the impact of each technique on the proposed architecture, an ablation study is presented in this section.
First, the LITE is stripped down from the three used techniques: dilation, multiplexing and custom filters.
Second, given that for the multiplexing convolutions performed in the first layer there are a total of three layers with $n$ filters, the Striped-LITE learns a total of $3n$ filters for the first layer.
The rest of the architecture is the same using DWSC without dilation.
We then add each boosting technique separately on the stripped LITE model and evaluate its performance.
The results of this ablation study are visualized in Figure~\ref{fig:ablation_LITE_results} in the form of a Heat Map.
Each cell of the Heat Map contains the Win Tie Loss count when evaluating the test accuracy on the UCR archive.
The addition of the $P$-Value statistics is presented in each cell in order to assess the significance in difference of performance.
The $P$-Value is emphasized in \textbf{bold} when it is lower than the specified threshold ($0.05$).
The colors of the Heat Map represent the difference in the average accuracy.

Results show that adding custom filters in the first layer as well as using multiplexing convolution in the first layer significantly boosts the performance.
The colors of the Heat Map indicates that adding the custom filters has also a positive impact on the average accuracy though it can add some parameters.
This is not the case for the multiplexing convolutions. We believe that this small average impact ($0.34\%$ overall) is not as important as the positive one. This is due to the fact that multiplexing reduces the number of parameters and wins over the majority of the datasets significantly.
The addition of the dilated convolution is shown to not have a statistical significance on the performance ($P$-Value $>0.05$). However, the average difference in accuracy shows that most of the times, using the dilated convolutions can improve the results. Given that dilation does not add more parameters and on average it boosts the performance, we keep it in the LITE architecture.
This is due to the fact that dilation increases the receptive field, so for large datasets this can be a boosting feature. The reason why sometimes Dilation can have a negative effect is because some of the datasets in the UCR archive do not require a large receptive field. 
Altogether, the LITE model (with the boosting techniques) will have less parameters compared to the striped LITE, while preserving performance compared to state-of-the-art models.
The decrease in number of parameters when using all the boosting techniques together comes from the fact that multiplexing removes more parameters than the custom filters adds.
Lastly, Figure~\ref{fig:ablation_LITE_results} shows the average rank of the models, such as in the CD Diagram explained in Section~\ref{sec:acc-perf}. The average rank of the Add-Custom-Filters is the lowest, while the Striped-LITE has the highest rank. Therefore, the worst model between the four presented in the Heat Map is the LITE without any boosting techniques.

\subsubsection{Impact of DWSC}
\begin{figure}[!ht]
    \centering
    \includegraphics[width=0.55\linewidth]{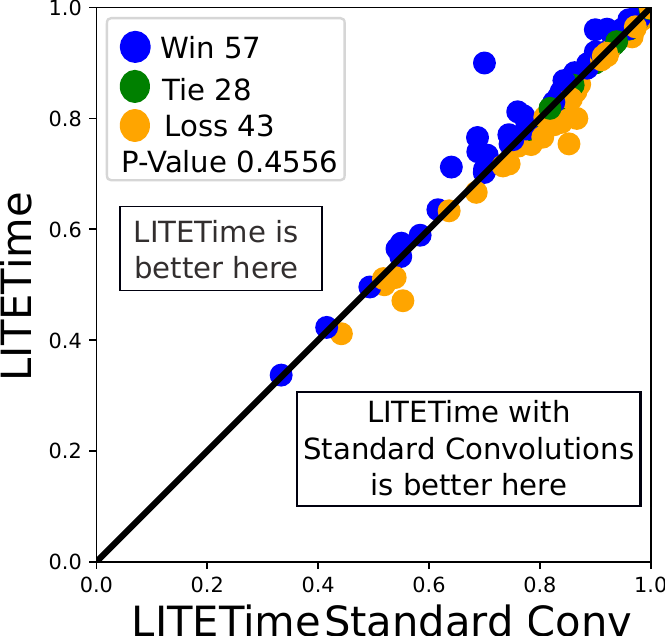}
    \caption{One-vs-one comparison between LITETime and LITETime with Standard convolutions over the 128 datasets of the UCR archive.}
    \label{fig:res_LITEtime_separable_conv}
\end{figure}

To show the effect of DWSC as well, we replace them by standard convolutions followed by a BottleNeck. To get a non-noisy comparison, we used ensembles. 
Note that the usage of the ensemble technique is necessary in this case given that by removing the DWSC, the difference in number of parameters becomes very high (LITE has almost $11\%$ of the compared model's number of parameters (the compared model has around $85,000$ parameters).
In Figure~\ref{fig:res_LITEtime_separable_conv}, the one-vs-one comparison between LITETime and LITETime with Standard Convolutions is presented. Results show that the usage of DWSC does not have an effect on the performance because the $P$-Value is high $0.4556$.
This means that the difference in performance is not statistically significantly different with less parameters mainly due to the usage of the DWSC.

\subsubsection{Number of LITE Models in the Ensemble}
\begin{figure}
    \centering
    \includegraphics[width=0.5\textwidth]{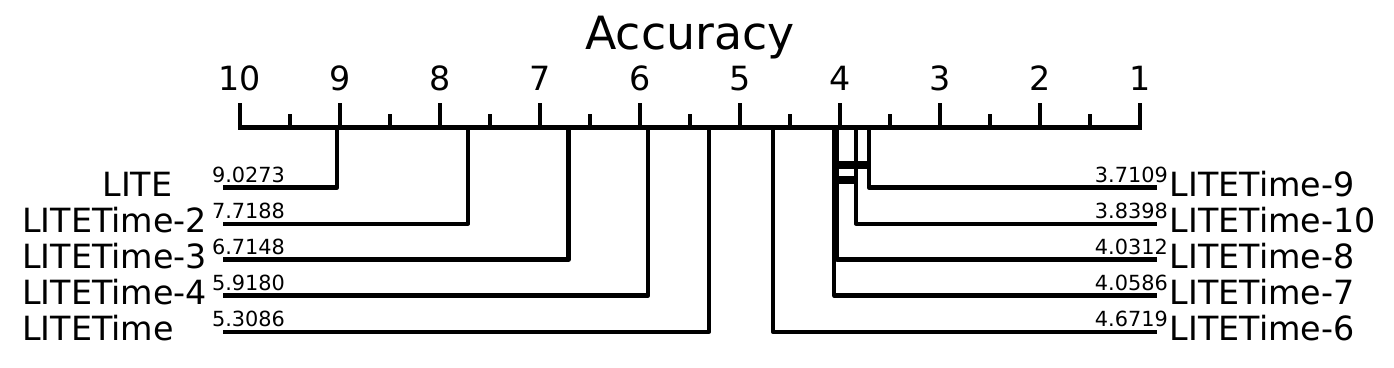}
    \caption{
    A Crtical Difference diagram showcasing the comparison of performance of LITETime when more or less LITE models are used in the ensemble.
    }
    \label{fig:ensemble_study}
\end{figure}

Given that LITETime is an ensemble of multiple LITE models, we stick to using five LITE models in LITETime in order to be fairly comparable to InceptionTime (ensemble of five Inception models).
It has been experimentally shown in~\cite{ismail2020inceptiontime} that no significant difference of performance can be found on the UCR archive when more than five Inception models are considered in the ensemble InceptionTime.
However, in the case of LITE, given its small architecture, its more bound to produce more variance and be less robust than Inception.
For this reason, we believe that the number of LITE models can be higher in the ensemble LITETime.
To experimentally prove our hypothesis, we trained ten different LITE models on all the UCR archive.
Ensembles of $1,2,3,\ldots,10$ are constructed by averaging over all possible ensemble combinations.
For instance, if the goal is to construct LITETime$-3$ (ensemble of three LITE models), we constructed the ensemble of all possible combinations of three LITE models from the pool of ten trained models.
We present the results on the CD diagram in Figure~\ref{fig:ensemble_study}.

\begin{figure}
    \centering
\includegraphics[width=0.45\textwidth]{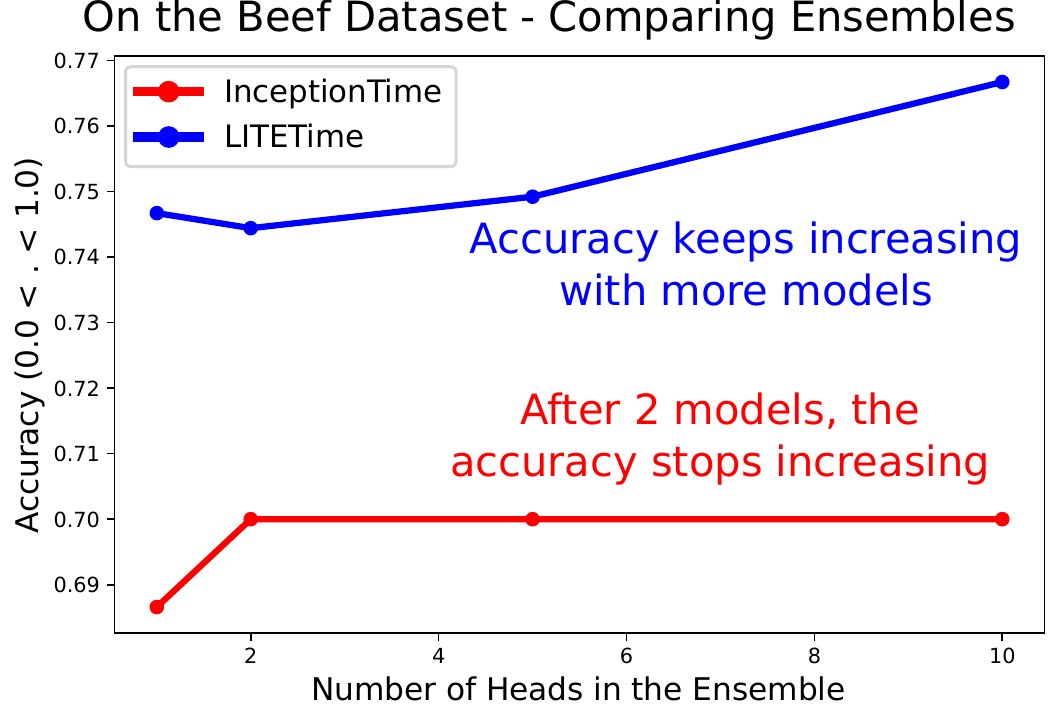}
    \caption{
    A Comparison on the Beef dataset of the UCR archive between the ensemble of LITE and Inception models.
    The $x$ axis represents the number of models used in each ensemble and the $y$ axis the performance of the ensemble on the test set of the Beef dataset.
    }
    \label{fig:ensemble_on_beef_vary}
\end{figure}

It can be seen from Figure~\ref{fig:ensemble_study} that in the case of LITE, LITETime$-5$ (ensemble of five LITE models) is not the limit that we can achieve, instead it is LITETime$-7$.
This is due to the fact that LITE is almost $42$ times smaller than Inception.
The ability to achieve this comes with the advantage of enhancing accuracy whilst LITETime exhibits a considerably smaller complexity compared to InceptionTime.
LITETime$-5$ with five models has almost $2.34\%$ of InceptionTime's trainable parameters but LITETime$-7$ increases this percentage only to $3.27$.
This can be seen with an example on the Beef dataset of the UCR archive in Figure~\ref{fig:ensemble_on_beef_vary}.
In this figure, we plot how the number of models in the case of LITETime and InceptionTime changes the performance on unseen data.

\subsection{Accuracy Performance on the UEA Archive}\label{sec:results_uea}
\begin{figure}
    \centering
    \includegraphics[width=0.5\textwidth]{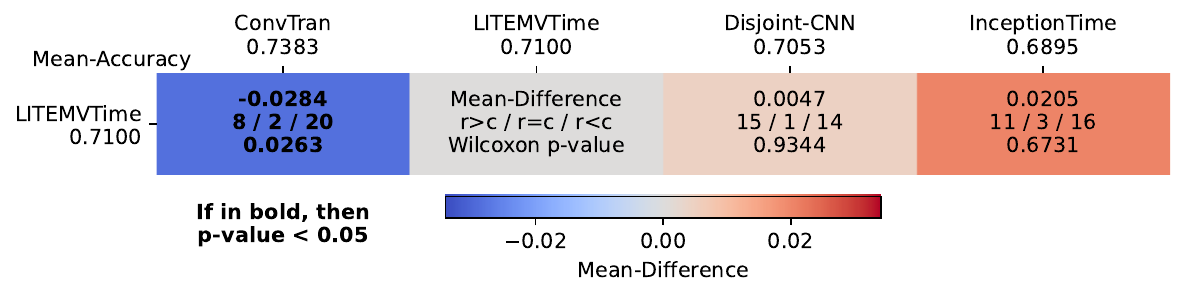}
    \caption{A Multi-Comparison Matrix (MCM) showcasing the performance of LITEMVTime, InceptionTime, Disjoint-CNN and ConvTran on the 30 datasets of the UEA archive.}
    \label{fig:mcm_uea}
\end{figure}

In Table~\ref{tab:results_uea_acc}, we present the accuracy performance of LITEMVTime, LITETime and five different competitors: ConvTran, InceptionTime, Disjoint-CNN, FCN and ResNet.
The accuracies of all five competitors are reported from the ConvTran paper~\cite{foumani2023improving}.
For the LITEMVTime model, we can see that when it wins over other models, a significant gap is observed between the accuracies.
This can be clearly seen as well in the MCM plot of LITEMVTime with the rest of the models in Figure~\ref{fig:mcm_uea}.
It can be seen that LITEMVTime comes second on the average performance, outperforming Disjoint-CNN and InceptionTime.
LITEMVTime still does not win on more than seven datasets compared to ConvTran, but looking deeper into Table~\ref{tab:results_uea_acc}, it can be seen that LITEMVTime can win with a large margin of accuracy.
For instance, for the EigenWorms dataset, the highest accuracy is reported by ConvTran with a value of $59.34\%$, however LITEMVTime produces an accuracy of $93.89\%$ and LITETime with an accuracy of $95.42\%$.

\begin{table*}[]
\centering
\caption{Accuracy performance in $\%$ of LITEMVTime (LMVT), LITETime (LT), ConvTran (CT), InceptionTime (IT), Disjoint-CNN (D-CNN), FCN and ResNet on $30$ datasets of the UEA archive. The datasets are ordered by their average number of training samples per class. The accuracy of the best model for each dataset is presented in bold and of the second best is underlined.}
\label{tab:results_uea_acc}
\begin{tabular}{|c|c|c|c|c|c|c|c|c|}
\hline
Dataset &
  \begin{tabular}[c]{@{}c@{}}Train Size\\ per Class\end{tabular} &
  LMVT &
  LT &
  CT &
  IT &
  D-CNN &
  FCN &
  ResNet \\ \hline
FaceDetection             & 2945 & 61.01          & {\ul 62.37}    & \textbf{67.22} & 58.85          & 56.65          & 50.37          & 59.48          \\ \hline
InsectWingbeat            & 2500 & 61.72          & 39.79          & \textbf{71.32} & {\ul 69.56}    & 63.08          & 60.04          & 65.00          \\ \hline
PenDigits                 & 750  & \textbf{98.86} & {\ul 98.83}    & 98.71          & 97.97          & 97.08          & 98.57          & 97.71          \\ \hline
SpokenArabicDigits        & 660  & 98.59          & {\ul 98.77}    & \textbf{99.45} & 98.72          & 98.59          & 98.36          & 98.32          \\ \hline
LSST                      & 176  & \textbf{66.42} & {\ul 62.85}    & 61.56          & 44.56          & 55.59          & 56.16          & 57.25          \\ \hline
FingerMovements           & 158  & \textbf{56.00} & 44.00          & \textbf{56.00} & \textbf{56.00} & {\ul 54.00}    & 53.00          & {\ul 54.00}    \\ \hline
MotorImagery              & 139  & 53.00          & 51.00          & \textbf{56.00} & 53.00          & 49.00          & {\ul 55.00}    & 52.00          \\ \hline
SelfRegulationSCP1        & 134  & 73.04          & 75.09          & \textbf{91.80} & 86.34          & {\ul 88.39}    & 78.16          & 83.62          \\ \hline
Heartbeat                 & 102  & 61.46          & 67.80          & \textbf{78.53} & 62.48          & 71.70          & 67.80          & {\ul 72.68}    \\ \hline
SelfRegulationSCP2        & 100  & {\ul 55.00}    & 53.89          & \textbf{58.33} & 47.22          & 51.66          & 46.67          & 50.00          \\ \hline
PhonemeSpectra            & 85   & 15.81          & 17.45          & \textbf{30.62} & 15.86          & {\ul 28.21}    & 15.99          & 15.96          \\ \hline
CharacterTrajectories     & 72   & \textbf{99.58} & {\ul 99.51}    & 99.22          & 98.81          & 99.45          & 98.68          & 99.45          \\ \hline
EthanolConcentration      & 66   & \textbf{69.20} & {\ul 67.30}    & 36.12          & 34.89          & 27.75          & 32.32          & 31.55          \\ \hline
HandMovementDirection     & 40   & 35.14          & 21.62          & {\ul 40.54}    & 37.83          & \textbf{54.05} & 29.73          & 28.38          \\ \hline
PEMS-SF                   & 39   & 79.19          & 82.66          & 82.84          & \textbf{89.01} & \textbf{89.01} & {\ul 83.24}    & 73.99          \\ \hline
RacketSports              & 38   & 73.68          & 78.29          & \textbf{86.18} & 82.23          & {\ul 83.55}    & 82.23          & 82.23          \\ \hline
Epilepsy                  & 35   & \textbf{99.28} & {\ul 98.55}    & {\ul 98.55}    & \textbf{99.28} & 88.98          & \textbf{99.28} & \textbf{99.28} \\ \hline
JapaneseVowels            & 30   & 96.49          & 97.30          & \textbf{98.91} & 97.02          & {\ul 97.56}    & 97.30          & 91.35          \\ \hline
NATOPS                    & 30   & 90.00          & 88.89          & \textbf{94.44} & 91.66          & {\ul 92.77}    & 87.78          & 89.44          \\ \hline
EigenWorms                & 26   & {\ul 93.89}    & \textbf{95.42} & 59.34          & 52.67          & 59.34          & 41.98          & 41.98          \\ \hline
UWaveGestureLibrary       & 15   & 84.68          & 85.00          & {\ul 89.06}    & \textbf{90.93} & {\ul 89.06}    & 85.00          & 85.00          \\ \hline
Libras                    & 12   & {\ul 89.44}    & 87.78          & \textbf{92.77} & 87.22          & 85.77          & 85.00          & 83.89          \\ \hline
ArticularyWordRecognition & 11   & 97.33          & 97.67          & {\ul 98.33}    & \textbf{98.66} & \textbf{98.66} & 98.00          & 98.00          \\ \hline
BasicMotions &
  10 &
  \textbf{100.0} &
  {\ul 95.00} &
  \textbf{100.0} &
  \textbf{100.0} &
  \textbf{100.0} &
  \textbf{100.0} &
  \textbf{100.0} \\ \hline
DuckDuckGeese             & 10   & 18.00          & 24.00          & \textbf{62.00} & 36.00          & {\ul 50.00}    & 36.00          & 24.00          \\ \hline
Cricket                   & 9    & {\ul 98.61}    & 97.22          & \textbf{100.0} & {\ul 98.61}    & 97.72          & 93.06          & 97.22          \\ \hline
Handwriting               & 6    & \textbf{40.00} & 36.82          & 37.52          & 30.11          & 23.72          & {\ul 37.60}    & 18.00          \\ \hline
ERing                     & 6    & 84.44          & 89.63          & \textbf{96.29} & {\ul 92.96}    & 91.11          & 90.37          & {\ul 92.96}    \\ \hline
AtrialFibrillation        & 5    & 13.33          & 06.67          & \textbf{40.00} & 20.00          & \textbf{40.00} & {\ul 33.33}    & {\ul 33.33}    \\ \hline
StandWalkJump             & 4    & \textbf{66.67} & {\ul 60.00}    & 33.33          & 40.00          & 33.33          & 40.00          & 40.00          \\ \hline \hline
\textbf{Average}          & -    & {\ul 70.99} & 69.37    & \textbf{73.83}          & 68.95          & 70.52          & 67.70          & 67.20          \\ \hline
\end{tabular}
\end{table*}

\begin{figure}
    \centering
    \includegraphics[width=0.5\textwidth]{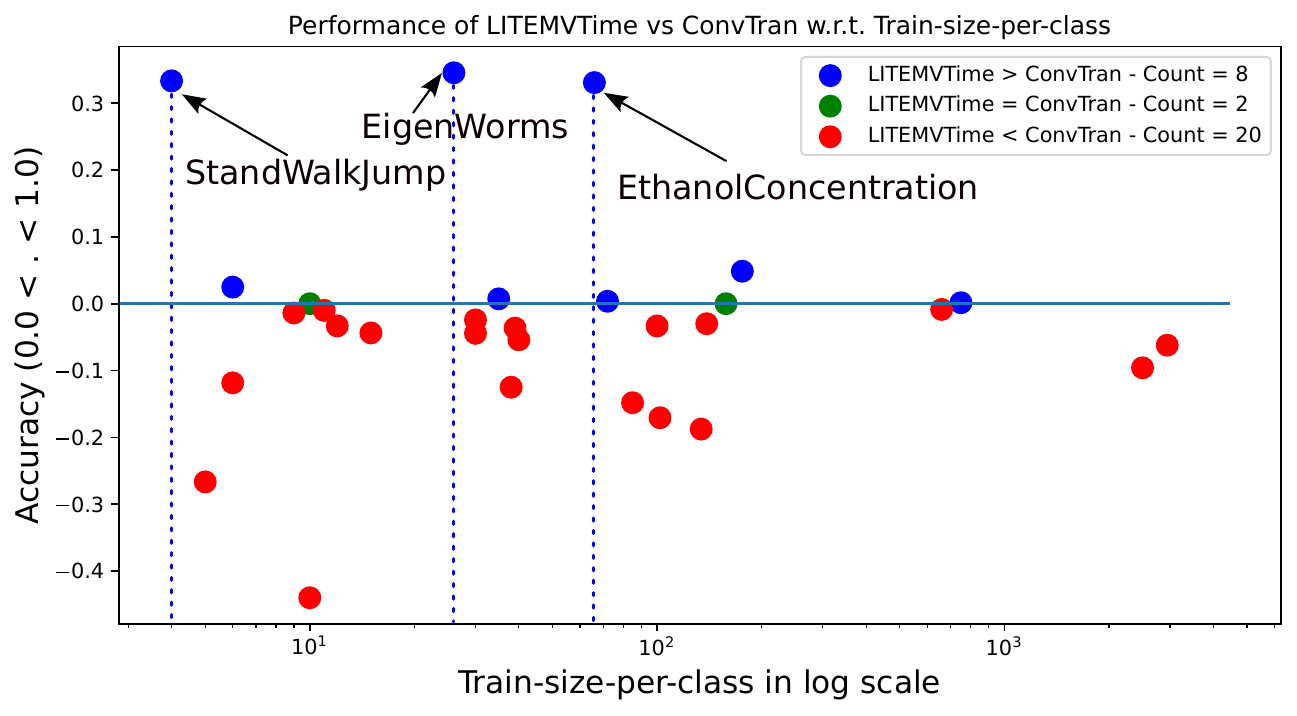}
    \caption{A one-vs-one comparison between CongTran and LITEMVTime with respect to the average training size per class. 
    Each point represents one dataset with its $x$ coordinate being the average training size per class (in $\log$ scale) and its $y$ coordinate being the difference between the accuracy of LITEMVTime and ConvTran.
    A positive value on the $y$ axis indicates a win for LITEMVTime.}
    \label{fig:litemv_convTran}
\end{figure}

To study in more details the cases where LITEMVTime works better than ConvTran with a significant margin, we present in Figure~\ref{fig:litemv_convTran} the difference in performance between these two models in function of the training samples per class.
This study is done to make sure if there are any common information about the cases where LITEMVTime works much better than ConvTran.
From Figure~\ref{fig:litemv_convTran}, we can see the most significant differences in performance are in the case of three datasets: StandWalkJump, EigenWorms and EthanolConcentration.
It can be seen that these datasets do not have the same range in the number of training examples.
For instance, StandWalkJump is considered to have a small training set with $4$ training examples per class and $12$ training examples in total.
However, the EthanolConcentration dataset has $66$ training examples per class and $261$ training samples in total.

\begin{figure}
    \centering
    \includegraphics[width=0.5\textwidth]{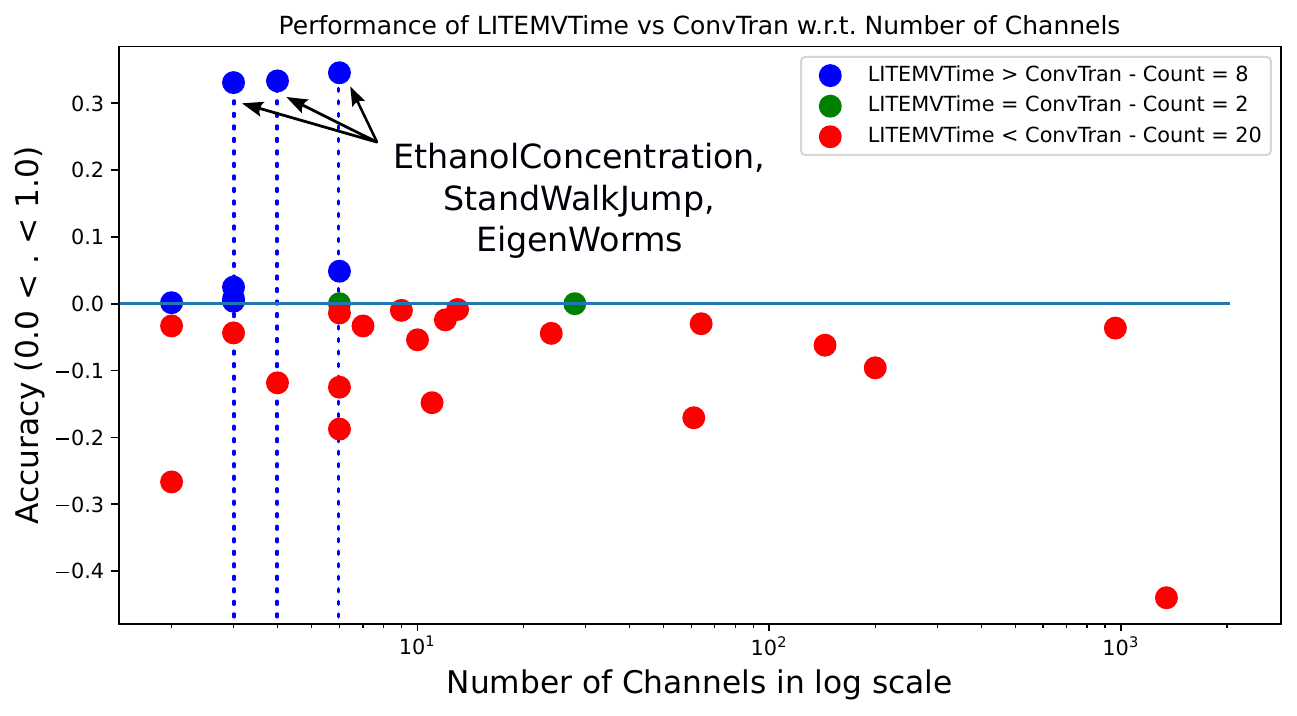}
    \caption{A one-vs-one comparison between CongTran and LITEMVTime with respect to the number of channels in the input multivariate time series.
    Each point represents one dataset with its $x$ coordinate being the number of channels (in $\log$ scale) and its $y$ coordinate being the difference between the accuracy of LITEMVTime and ConvTran.
    A positive value on the $y$ axis indicates a win for LITEMVTime.}
    \label{fig:litemv_convTran_dims}
\end{figure}

Another approach to study the reason why these datasets would work better than ConvTran when using LITEMVTime is to dig into the number of dimensions, taking advantage that this is the case of multivariate time series data.
In Figure~\ref{fig:litemv_convTran_dims}, we present the same information as in Figure~\ref{fig:litemv_convTran} but in function of the number of dimensions of the MTS datasets.
It can be seen that the same three datasets have a small number of dimensions.
It can be also seen that ConvTran always wins when the number of dimensions increases.

\subsection{Experiments on Human Motion Rehabilitation}
To assess the performance of LITEMVTime in a real life application, we consider the Human Rehabilitation domain.
For this domain, the data is extracted from patients into the form of 3D skeleton based sequences.
A human expert has to asses for each patient if they are doing a movement (rehabilitation exercise) correctly or not.
This is a very important application and is highly correlated with this work given that this type of data can be regarded as Multivariate Time Series.
For each sequence recorded, we have a specific number of joints in a three dimensional space. 
In other words, each sequence is a multivariate time series with $3*J$ channels where $J$ is the number of human joints.

\begin{figure*}
    \centering
    \includegraphics[width=\textwidth]{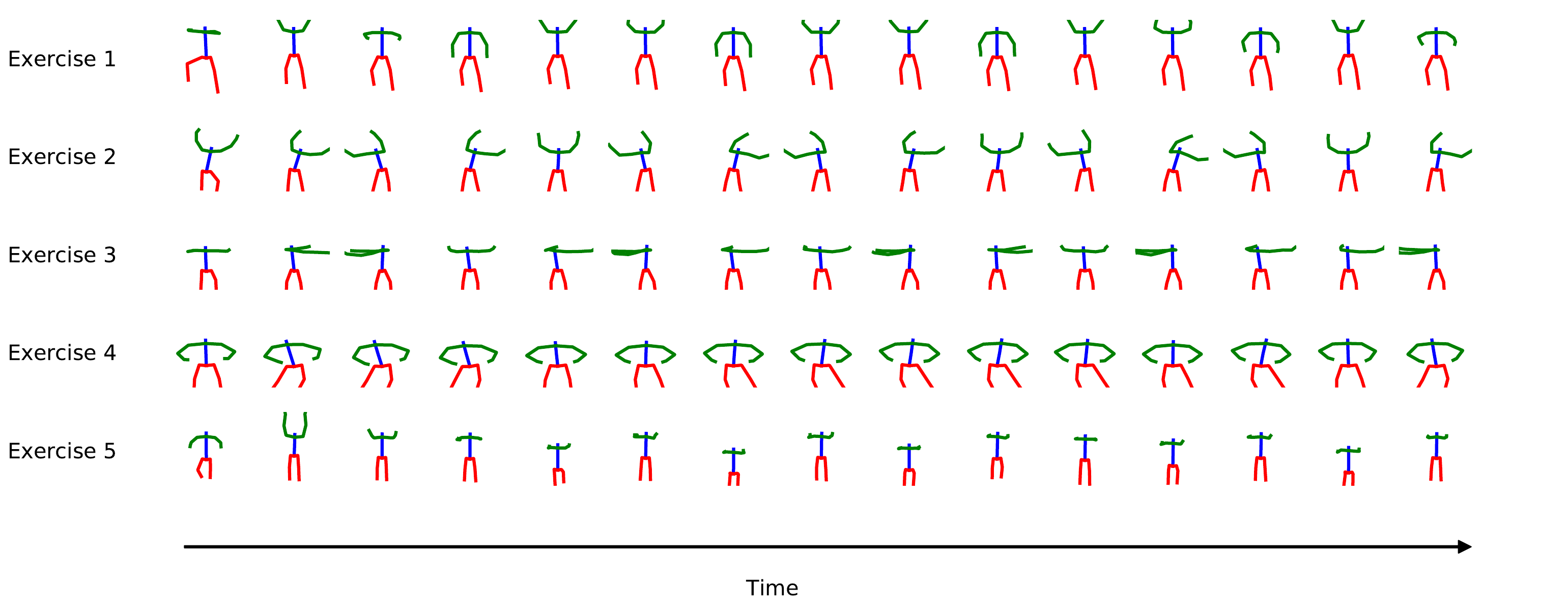}
    \caption{One sample from each of the rehabilitation exercises of the Kimore dataset.
    Each sample showcases how different parts of the body change positions through time.
    The skeletons are color coded following the five different body parts: left and right legs (in \colorbox{redF2}{red}), left and right arms (in \colorbox{greenF2}{green}) and spine (in \colorbox{blueF2}{blue}).
    }
    \label{fig:kimore_examples}
\end{figure*}

\begin{figure*}
    \centering
    \includegraphics[width=\textwidth]{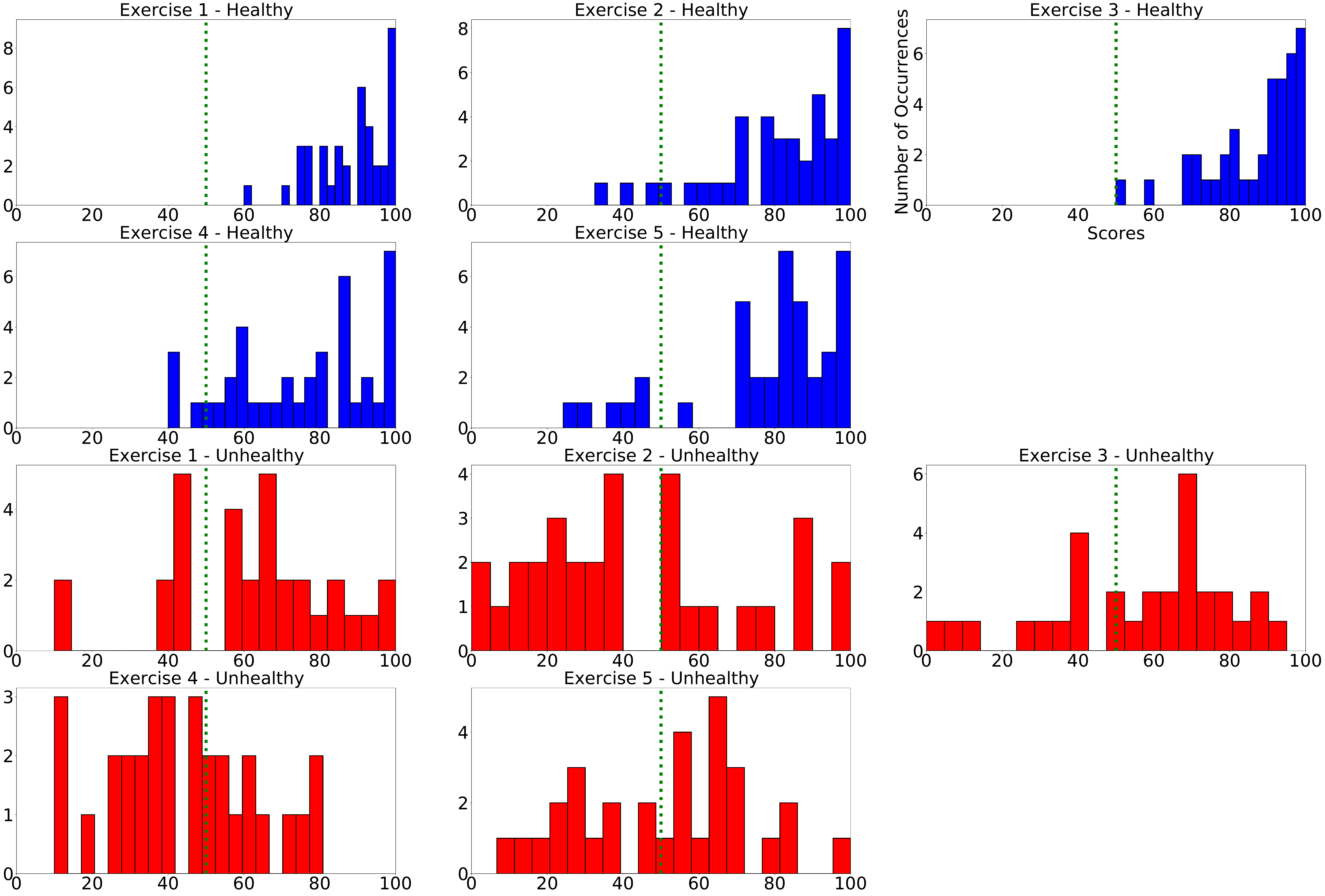}
    \caption{
    The distribution of the scores given by experts to healthy (in \colorbox{blueF2}{blue}) and unhealthy (in \colorbox{redF2}{red}) patients when performing each of the five different exercises.
    The threshold set to discretize these scores is chosen to be the middle point (line in green) posed at $50$.
    }
    \label{fig:kimore_scores_dist}
\end{figure*}

For this experiment, we used the Kimore dataset~\cite{capecci2019kimore}.
This dataset contains recorded sequences in the form of videos of patients performing rehabilitation exercises, which are then transformed into numerical MTS using kinect v2~\cite{lun2015survey_kinect}.
This dataset contains both healthy and unhealthy subjects performing five different rehabilitation exercises.
A human expert then assesses the quality of the performed exercise by providing a score between $0$ and $100$, from bad to good respectively.
A visualization of the distribution of these scores for each exercise is presented in Figure~\ref{fig:kimore_scores_dist} in the case of healthy and unhealthy subjects.
It can be seen, for instance, that most cases of unhealthy subjects tend to have a low score and a high score in the case of healthy subjects.
Given that this is a regression dataset, we re-orient the task to learn a deep learning model to evaluate the performance of a subject regardless of being healthy or not.
Instead, the evaluation is done as following:
\begin{itemize}
    \item if the subject's score is less than $50$, the exercise is considered to be badly performed;
    \item if the subject's score is higher than $50$, the exercise is considered to be performed well.
\end{itemize}
Some examples of the five exercises are presented in Figure~\ref{fig:kimore_examples}.

Each recorded sequence has a different length, which should be fixed to one common length in order to use deep learning models.
We resampled~\footnote{https://docs.scipy.org/doc/scipy} all the samples in the data into the average length, which is $748$.
The sequences contain $18$ human joints with each in a 3D space.
The dataset is made of $71$ examples per exercise, where each sequence is a recording of one subject.
For this reason, we can randomly split for each exercise the dataset into a $80\%-20\%$ train test sets.
In order to make sure samples of each class exist in the train test split, we make sure to stratify when randomly splitting the train/test sets with the labels (good/bad).

All samples are $z$-normalized in order to have a zero mean and unit standard variation.
In Table~\ref{tab:kimore_results}, we present the accuracy performance for each of the five exercises using the three deep learning models from the literature: FCN, ResNet and InceptionTime, our proposed architecture LITEMVTime, and a baseline classifier 1-Nearest Neighbour-Dynamic Time Warping (1-NN-DTW)~\cite{middlehurst2023bake,bagnall2017great}.
The experiments using 1-NN-DTW are conducted using \textit{aeon} Python pacakge~\cite{middlehurst2024aeon}.
It can be concluded from the table that LITEMVTime is the best performing model for this task compared to the competitors following both the average performance and average rank over all exercises.
This results show that a small deep learning model such as LITEMVTime can be used to classify if a patient is healthy or not given a recorded sequence of their exercise.

\begin{table*}[]
\caption{
Accuracy of the baseline, 1-NN-DTW, three state-of-the-art deep learning models, FCN ResNet and InceptionTime compared to \textbf{ours} LITEMVTime on the Kimore human rehabilitation exercise.
We present for each of the five exercises the accuracy of the models on the test unseen split.}
\label{tab:kimore_results}
\begin{tabular}{c|ccccc}
Kimore Exercise & \textbf{1-NN-DTW} & \textbf{FCN} & \textbf{ResNet} & \textbf{InceptionTime} & \textbf{LITEMVTime} \\ \hline
\textbf{Exercise 1}       & 60.00          & 84.00          & {\ul 85.33} & 78.67          & \textbf{86.67} \\
\textbf{Exercise 2}       & 46.67          & 72.00          & 69.33       & {\ul 78.67}    & \textbf{80.00} \\
\textbf{Exercise 3}       & 86.67          & \textbf{92.00} & 86.67       & {\ul 88.00}    & 86.67          \\
\textbf{Exercise 4}       & \textbf{66.67} & {\ul 65.33}    & 60.00       & 57.33          & \textbf{66.67} \\
\textbf{Exercise 5}       & 73.33          & 66.67          & {\ul 81.33} & \textbf{84.00} & 80.00          \\ \hline
\textbf{Average Accuracy} & 66.67          & 76             & 76.53       & {\ul 77.33}    & \textbf{80.00} \\ \hline
\textbf{Average Rank}     & 3.6            & 2.8            & 2.8         & {\ul 2.6}      & \textbf{1.8} 
\end{tabular}
\end{table*}

\subsection{Explainability: Class Activation Map}
Having a good performing deep learning model is important. However it is important to also be able to understand the decision making process of a deep learning model.
This field has been significantly targeted in the last decade and in the domain of TSC for the last five years.
Class Activation Map (CAM) is an explainability technique for deep CNNs, which helps interpreting the decision of CNNs as if they were black box models.
It was first introduced in~\cite{zhou2016learning} for image datasets and got first adapted to time series data in~\cite{fcn_resnet}.
It is important to note that the usage of CAM explainability necessities the usage of a global representative layer before the softmax classification layer.
An example of such layers is the Global Average Pooling (GAP) used in FCN, ResNet, Inception, LITE and LITEMV.
The outcome of a CAM in the case of TSC is a univariate time series.
Each time stamp represents the importance of this time stamp in the input time series that lead to a specific decision making.

For the mathematical setup of CAM, we defined:
\begin{itemize}
    \item the output of the last convolution layer as $\textbf{O}(t)$, which is a multivariate time series with $M$ variables, $M$ being the number of filters of this layer;
    In other words, $\textbf{O}_m(t)$ is the output univariate time series of filter $m$, where $m \in \{0,1,\ldots,M-1\}$;
    \item $\textbf{w}^{c}=\{w_{0}^{c},w_{1}^{c},\ldots,w_{M-1}^{c}\}$ as the weights vector connecting the GAP output to the neuron of the winning class (the class with the highest probability value)
\end{itemize}
The CAM output can be defined as follows:
\begin{equation}
    CAM(t) = \sum_{m=0}^{M-1} w_m^{c}.\textbf{O}_m(t)
\end{equation}
This is followed by a $\min-\max$ normalization on the CAM output.
For two given time stamps, the one with the highest CAM score contributed more in the decision making of the black box deep learning model.

\begin{figure*}
    \centering
    \includegraphics[width=\textwidth]{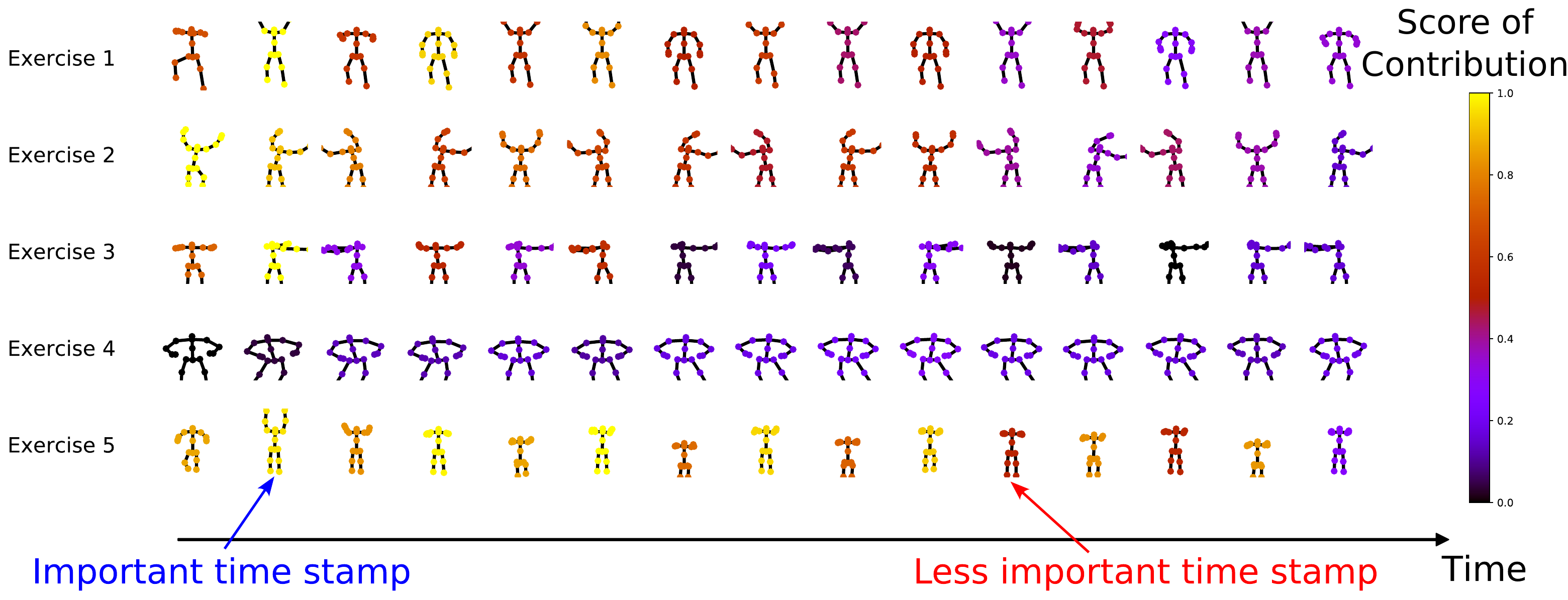}
    \caption{
    Interpretability of the LITEMV model using the Class Activation Map (CAM) on the feature of the last DWSC layer.
    The colorbar values represent the normalized (between $0$ and $1$) scores of the CAM.
    Five samples from each of the five exercises are presented with the CAM scores on different time stamps.
    A higher CAM score indicates the importance of a time stamp for the decision making of LITEMV.
    }
    \label{fig:kimore_cam}
\end{figure*}

In this work, we use the CAM explainability technique to better understand the decision making of the LITEMV model on the human rehabilitation dataset Kimore.
We use five different examples from each exercise, and produce the CAM output using a LITEMV model trained to solve the classification task of each exercise (see Figure~\ref{fig:kimore_cam}).
Given that in the case of human skeleton based data, the input is a multivariate time series, we suppose that the produced CAM values represent the temporal axis.
In other words, for each time stamp, one CAM score represents the contribution of the current time pose (skeleton) to the classification.

\begin{figure*}
    \centering
    \includegraphics[width=\textwidth]{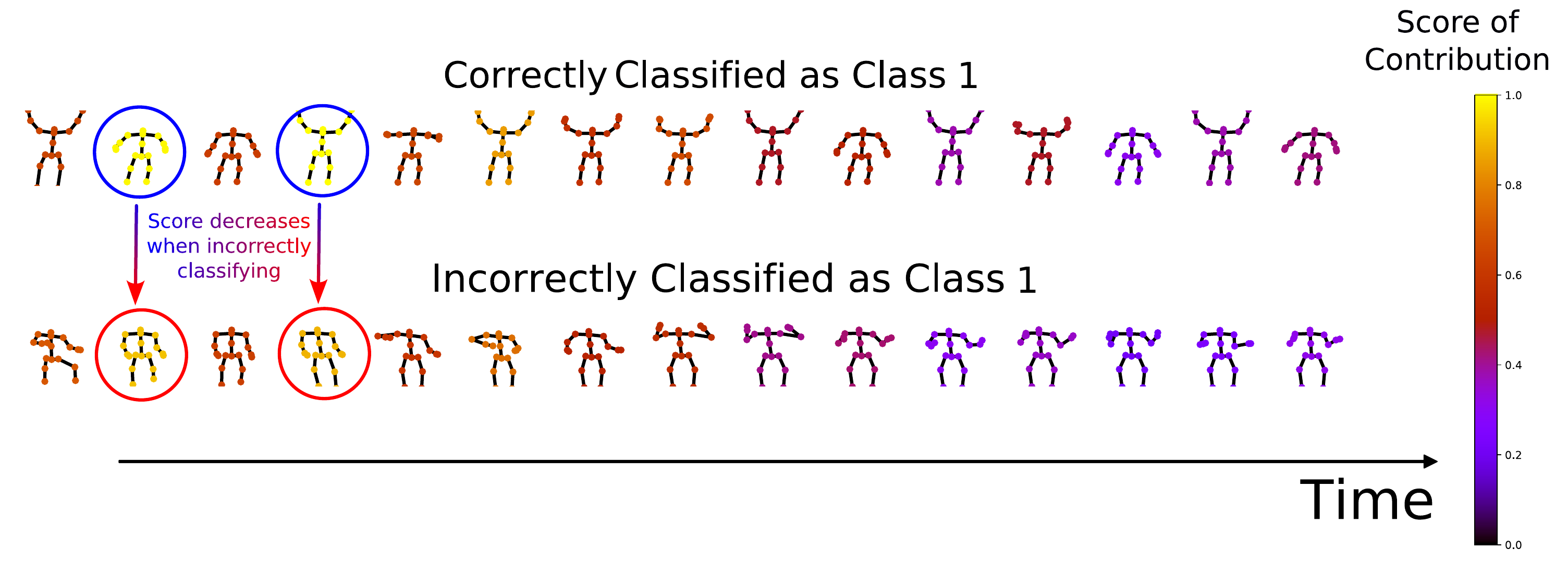}
    \caption{
    Interpretability of the LITEMV model using the Class Activation Map (CAM) on the feature of the last DWSC layer.
    Two samples from the test split of the same exercise are presented, the first (top) having a ground truth of class 1, and the second (bottom) having a ground truth of class 0.
    LITEMV correctly classifies the first sample but incorrectly the second.
    It can be seen that the important time stamp in the case of the correctly classified sample has higher color intensity, so higher CAM score, compared to the same time stamp from the incorrectly classified sample.
    }
    \label{fig:kimore_cam_diff}
\end{figure*}

To study more the changes in the CAM scores depending on the state of correctly classifying a sample, we present in Figure~\ref{fig:kimore_cam_diff} two CAM explanations on two samples of the same exercise.
The first sample is correctly classified as class 1 ($score > 50$) and the second is incorrectly classified as class 1 as it should be class 0 ($score < 50$).
It can be seen that the first sample (top) has higher intensity in the CAM colors (so in the CAM scores), than the second sample (bottom).
This indicates that the important time stamps for correctly classifying as class 1 will have lower scores, when the ground truth is actually class 0.
This is due to the fact the CAM scores consider the weights of the winning class in the classification layer and not the weights of the ground truth.

\subsection{Limitations of LITE and LITEMV}
LITE and LITEMV have low complexity when compared to other architectures; so, one potential limitation can occur when using  these methods for handling big data. For instance, LITE and LITEMV could fall short when a training set including millions of data is used. 
This shortcoming could be solved by increasing the number of filters operating in the DWSC layers: in fact, given the efficient way DWSC's apply convolutions, this would result in a slight increase in the cost.

A second limitation that affects all the architectures detailed in this work, is related to the length of the time series samples. 
This can be circumvented by increasing the CNN's Receptive Field (RF), which is the length of input visibility of the CNN at the last layer.
Let us consider a CNN with $L$ convolution layers, of kernel size $K_i$ and dilation rate $d_i$, where $i \in \{0,1,\ldots,L-1\}$.
The Receptive Field~\cite{liu2018understanding, luo2016understanding} of this model is computed as:
\begin{center}
    $RF = 1+\sum_{d_i*K_i}$.
\end{center}

\noindent
This quantity varies from one CNN to another. For instance, in the case of FCN~\cite{fcn_resnet}, the RF is $14=1+7+4+2$.
This RF is small compared to the length of the time series in the UCR archive.
In the case of ResNet, the RF increases to $40$, and to $235$ for Inception.
In the case of LITE and LITEMV, the RF is $114$, which has been shown to be sufficient to achieve state-of-the-art performance on the UCR archive.
However, this RF value needs to be increased if a dataset includes a significantly longer time series compared to this value.
The RF value can be increased by either increasing the length of the filters, or by adding more layers to increase the model's depth. In usual CNN models, however, this is not an efficient approach given the fact it drastically increases the complexity of the network. This does not occurs for LITE and LITEMV, making them suitable for such a solution.

\section{Conclusions}\label{sec:conclusion}
In this paper, we addressed the Time Series Classification problem by reducing the number of parameters compared to existing deep learning approaches for Time Series Classification, while preserving performance of InceptionTime.
We presented a new architecture for Time Series Classification, LITE, and evaluated its performance on the UCR archive. 
LITE has only $2.34\%$ of InceptionTime's number of parameters.
This model is faster than the state-of-the-art in training and inference time. It consumes as well less CO2 and power, which is a topic we believe to be very important nowadays.
Results have illustrated that the usage of LITE allows us to achieve state-of-the-art results on the UCR archive.
Furthermore, the presented ablation study demonstrated the importance of the techniques used in LITE.
Furthermore, we also adapted LITE to handle multivariate time series data, LITEMV.
Experiments on the UEA multivariate TSC archive showed promising performance on some datasets compared to the state-of-the-art.
Finally, we showcased the utility of LITEMV in a real application, where human rehabilitation exercises are evaluated. In this context, we showed that LITEMVTime outperforms other models on the Kimore dataset.
We believe this work can represent a starting point for optimizing deep learning architectures in the time series domain. We believe that this study can address clustering, representation learning and generative models as well. In future work, we aim to tackle these others domains given
the impressive performance of LITE compared to ResNet and InceptionTime.

\bmhead{Acknowledgments}
This work was supported by the ANR DELEGATION
project (grant ANR-21-CE23-0014) of the French
Agence Nationale de la Recherche. The authors
would like to acknowledge the High Performance
Computing Center of the University of Strasbourg
for supporting this work by providing scientific sup-
port and access to computing resources. Part of the
computing resources were funded by the Equipex
Equip@Meso project (Programme Investissements
d’Avenir) and the CPER Alsacalcul/Big Data. The
authors would also like to thank the creators and
providers of the UCR, UEA Archives and the Kimore dataset.

\section*{Declarations}

\begin{itemize}
\item \textbf{Funding} This work was supported by the ANR DELEGATION
project (grant ANR-21-CE23-0014) of the French
Agence Nationale de la Recherche.
\item \textbf{Conflict of interest} The authors certify that they have no conflict of
interest in the subject matter or materials discussed in this manuscript.
\item \textbf{Availability of data and materials} All of the datasets used in this work are publicly available.
\item \textbf{Code availability} The source code is available on this github repository: \texttt{https://github.com/MSD-IRIMAS/LITE}
\item \textbf{Authors' contributions} Conceptualization: AIF; Methodology: AIF, MD, SB, JW and GF; Experiments: AIF, MD; Validation: MD, SB, JW and GF; Writing—original draft: AIF; Writing—review and editing: AIF, MD, SB, JW and GF; Funding acquisition: MD; all authors have read and agreed to the published version of the manuscript.
\end{itemize}

\bibliography{sn-bibliography}

\end{document}